\pdfoutput=1
\RequirePackage[svgnames]{xcolor}

\documentclass[11pt]{article}

\usepackage{ACL2023}

\usepackage{times}
\usepackage{latexsym}

\usepackage[draft,textsize=footnotesize,textwidth=15mm]{todonotes}
\usepackage[utf8]{inputenc}
\usepackage{inconsolata}
\usepackage{algorithm}
\usepackage{algpseudocode}
\usepackage{amsmath,amssymb}
\usepackage{soul}
\usepackage{tikz}

\tikzset{rndblock/.style={rounded corners,rectangle,draw,scale=0.8,outer sep=0pt}}

\newcommand{\tframed}[2][]{\tikz[baseline=(h.base)]\node[rndblock,#1] (h) {#2};}

\usetikzlibrary{shapes.geometric}
\usepackage{framed}
\usepackage{enumitem}
\newlist{RQ}{enumerate}{1}
\setlist[RQ]{label=\textbf{RQ\,\arabic*},ref={RQ\,\arabic*}}
\usepackage{comment}
\usepackage{natbib}
\usepackage{multibib}
\makeatletter
\usepackage{booktabs}
\usepackage[inkscapeformat=png]{svg}
\usepackage{graphicx}
\usepackage{caption}
\usepackage{subcaption}
\usepackage{tabularx}
\usepackage{soul}
\usepackage{float}
\usepackage{enumitem}
\usepackage{pifont}
\usepackage{arydshln}
\usepackage{lipsum}

\usepackage[many]{tcolorbox}

\newtcolorbox{defin}{colback=Teal!5!White,enhanced,title=Contributions,
	attach boxed title to top left={xshift=-4mm},boxrule=0pt,after skip=1cm,before skip=1cm,right skip=0cm,breakable,fonttitle=\bfseries,toprule=0pt,bottomrule=0pt,rightrule=0pt,leftrule=3pt,arc=0mm,skin=enhancedlast jigsaw,sharp corners,colframe=Teal!55!black,colbacktitle=Teal!55!black,boxed title style={
		frame code={ 
			\fill[Teal!25!black](frame.south west)--(frame.north west)--(frame.north east)--([xshift=3mm]frame.east)--(frame.south east)--cycle;
			\draw[line width=1mm,Teal!25!black]([xshift=2mm]frame.north east)--([xshift=5mm]frame.east)--([xshift=2mm]frame.south east);
			\draw[line width=1mm,Teal!25!black]([xshift=5mm]frame.north east)--([xshift=8mm]frame.east)--([xshift=5mm]frame.south east);
			\fill[Teal!25!black](frame.south west)--+(4mm,-2mm)--+(4mm,2mm)--cycle;
		}
	}
}

\usetikzlibrary{shapes.geometric, arrows}
\usetikzlibrary{decorations.markings}

\usepackage{fancybox}
\usepackage{xcolor}
\usepackage{hyperref}
 \definecolor{darkblue}{rgb}{0, 0, 0.5}
  \hypersetup{colorlinks=true, citecolor=darkblue, linkcolor=darkblue, urlcolor=darkblue}

\definecolor{vgreen}{HTML}{60A917}
\definecolor{vred}{HTML}{CE3A29}

\usepackage{xstring}
\usepackage{longtable}

\usepackage{tabularray}

\DefTblrTemplate{firsthead,middlehead,lasthead}{default}{
}
\DefTblrTemplate{firstfoot}{default}{
  \UseTblrTemplate{contfoot}{default}
  \UseTblrTemplate{caption}{default}
}
\DefTblrTemplate{middlefoot}{default}{
  \UseTblrTemplate{contfoot}{default}
  \UseTblrTemplate{capcont}{default}
}
\DefTblrTemplate{lastfoot}{default}{
  \UseTblrTemplate{note}{default}
  \UseTblrTemplate{remark}{default}
  \UseTblrTemplate{capcont}{default}
}

\newcolumntype{P}[1]{>{\centering\arraybackslash}p{#1}}
\usepackage{color}
\tcbuselibrary{skins}

\usepackage[export]{adjustbox} % for the valign option

\usepackage{setspace}
\usepackage[capitalise,nameinlink]{cleveref}

\crefname{section}{Sec.}{Sec.}

\usepackage{microtype}
\usepackage{hyperref}
\usepackage{graphicx}
\usepackage{comment}
\usepackage{amsmath}
\usepackage{amssymb}
\usepackage{algorithm}
\usepackage{algpseudocode}
\usepackage{colortbl}
\usepackage[export]{adjustbox} % for the valign option
\usepackage{varwidth}
\usepackage{enumitem}
\setlist{leftmargin=1mm}
\usepackage{pifont}
\usepackage{booktabs}
\usepackage{multirow}
\usepackage{subcaption}
\usepackage{resizegather}
\usepackage{breqn}
\usepackage[capitalise]{cleveref}
\usepackage{graphicx}
\usepackage{tikz}
\usetikzlibrary{shapes.geometric, arrows}
\usetikzlibrary{decorations.markings}
\usepackage{soul}
\usepackage{wrapfig,graphicx,lipsum}% http://ctan.org/pkg/{wrapfig,graphicx,lipsum}
\usepackage{extsizes}
\usepackage{cuted}
\usepackage{flushend}
\usepackage{float}
\usepackage{changepage,threeparttable}
\usepackage{setspace}
\usepackage{caption}
\usepackage{booktabs}
\usepackage{dblfloatfix} 
\usepackage{fixltx2e}

\usepackage{environ}

\newlength{\myl}
\expandafter\let\expandafter\origequation\csname equation*\endcsname
\expandafter\let\expandafter\endorigequation\csname endequation*\endcsname
\long\def\[#1\]{\begin{equation*}#1\end{equation*}}
\RenewEnviron{equation*}{
  \settowidth{\myl}{$\displaystyle\BODY$} % calculate width and save as \myl
  \origequation
    \ifdim\myl>\linewidth
      \resizebox{\linewidth}{!}{$\displaystyle\BODY$}% \myl > \linewidth
    \else
      \BODY % \myl <= \linewidth
    \fi
  \endorigequation
}

\makeatletter
\newcommand{\DrawLine}{%
  \begin{tikzpicture}
  \path[use as bounding box] (0,0) -- (\linewidth,0);
  \draw[color=blue!75!black,dashed,dash phase=.5pt]
        (0-\kvtcb@leftlower-\kvtcb@boxsep,0)--
        (\linewidth+\kvtcb@rightlower+\kvtcb@boxsep,0);
  \end{tikzpicture}%
  }
\makeatother

\newcommand*{\Scale}[2][4]{\scalebox{#1}{$#2$}}%
\newcommand\acil[1]{\todo[author=AC,color=blue!40,inline]{#1}}
\newcommand\acb[1]{\textcolor{blue}{#1}}

\usepackage{euscript}[mathcal]

\newcommand*{\affaddr}[1]{#1}
\newcommand*{\affmark}[1][*]{\textsuperscript{#1}}
\newcommand*{\email}[1]{\texttt{#1}}
\author{
Vipula Rawte\affmark[1]\thanks{\,\,\,Corresponding author.}, S.M Towhidul Islam Tonmoy\affmark[2], Krishnav Rajbangshi\affmark[3], \\ \bf Shravani Nag\affmark[4], \bf Aman Chadha\affmark[5,6]\thanks{\,\,\,Work does not relate to position at Amazon.}, \bf Amit Sheth\affmark[1], Amitava Das\affmark[1]  \\
\affaddr{\affmark[1]AI Institute, University of South Carolina, USA}\\
\affaddr{\affmark[2]Islamic University of Technology}\\
\affaddr{\affmark[3]National Institute of Technology, Silchar}\\
\affaddr{\affmark[4]Indira Gandhi Delhi Technical University for Women}\\
\affmark[5]Stanford University, USA, 
\affmark[6]Amazon AI, USA \\
\email{\{vrawte\}@mailbox.sc.edu}
}

\title{\includegraphics[width=8cm]{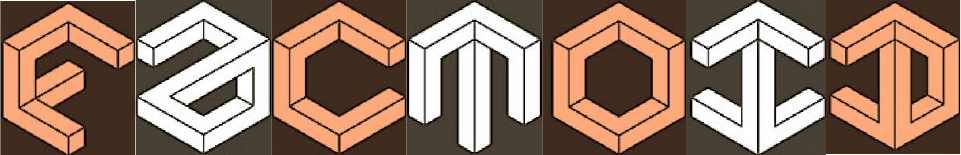}\\
\ul{FAC}tual en\ul{T}ailment f\ul{O}r halluc\ul{I}nation \ul{D}etection}

\let\oldtwocolumn\twocolumn
\renewcommand\twocolumn[1][]{%
    \oldtwocolumn[{#1}{
    \begin{center}
           \vspace{6mm}
           \includegraphics[width=0.9\textwidth]{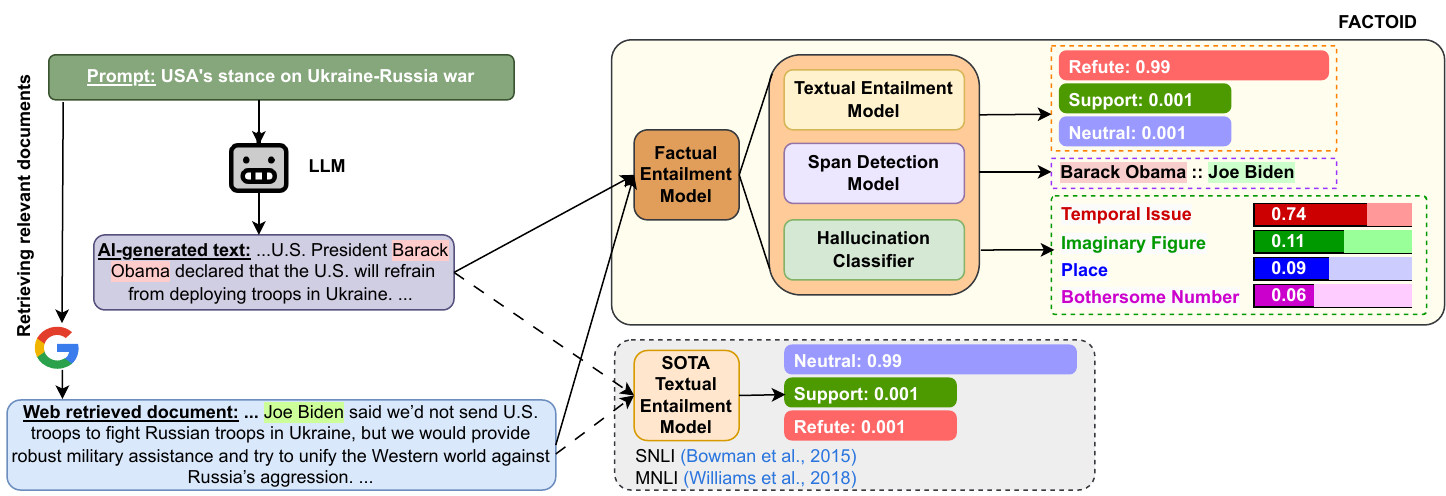}
           \captionof{figure}{An illustration of traditional Textual Entailment (TE) vs.our proposed Factual Entailment (FE). In part A (top), we emphasize the limitation of the TE method (trained on standard entailment tasks like SNLI \cite{bowman2015large} and/or MNLI \cite{williams-etal-2018-broad}, etc.) to recognize a case as a refute. In contrast, in part (B), the proposed Factual Entailment adopts a multitask learning approach that predicts an entailment score, hallucination type and the span of the entailment. FE therefore presents a novel approach to entailment that assists in identifying hallucinations. the retrieved document is a White House press release, could be see here: \href{https://www.whitehouse.gov/briefing-room/speeches-remarks/2022/04/28/remarks-by-president-biden-on-the-request-to-congress-for-additional-funding-to-support-ukraine/}{link}}
           \label{fig:fig1}\vspace{2mm}
        \end{center}
    }]
}

\begin{document}
\maketitle
\begin{abstract}
The widespread adoption of Large Language Models (LLMs) has facilitated numerous benefits and applications. However, among the various risks and challenges, hallucination is a significant concern. In response, Retrieval Augmented Generation (RAG) has emerged as a highly promising paradigm to improve LLM outputs by grounding them in factual information. RAG relies on textual entailment (TE) or similar methods to check if the text produced by LLMs is supported or contradicted, compared to retrieved documents. This paper argues that conventional TE methods are inadequate for spotting hallucinations in content generated by LLMs. For instance, consider a prompt about the ''\emph{USA's stance on the Ukraine war}''. The AI-generated text states, ``\emph{...U.S. President Barack Obama says the U.S. will not put troops in Ukraine...}'' However, during the Ukraine-Russia war, the U.S. president is Joe Biden, not Barack Obama, which contradicts factual reality. Moreover, current TE systems are unable to accurately annotate the given text and identify the exact portion that is contradicted. To address this challenge, this paper introduces a new type of TE called ``\ul{Factual Entailment (FE)}.'', aims to detect factual inaccuracies in content generated by LLMs while also highlighting the specific text segment that contradicts reality. We present $\mathcal{FACTOID}$ (FACTual enTAILment for hallucInation Detection), a benchmark dataset for FE. We propose a multi-task learning (MTL) framework for FE, incorporating state-of-the-art (SoTA) long text embeddings such as e5-mistral-7b-instruct, along with GPT-3, SpanBERT, and RoFormer. The proposed MTL architecture for FE achieves an avg. 40\% improvement in accuracy on the $\mathcal{FACTOID}$ benchmark compared to SoTA TE methods. As FE automatically detects hallucinations, we assessed 15 modern LLMs and ranked them using our proposed \emph{\ul{Auto Hallucination Vulnerability Index ($HVI_{auto}$)}}. This index quantifies and offers a comparative scale to evaluate and rank LLMs according to their likelihood of producing hallucinations. FACTOID dataset\footnote{\scriptsize\url{https://huggingface.co/datasets/aisafe/FACTOID}} and demo\footnote{\scriptsize\url{https://huggingface.co/spaces/aisafe/FACTOID}} are publicly available.

%Similarly, another example is that we said we’d not send U.S. troops to fight Russian troops in Ukraine, but we would provide robust military assistance and try to unify the Western world against Russia’s aggression \cite{White-house}. 
%Our objective is to automate the critical process of hallucination detection, which is increasingly essential for the effective integration of LLMs into various downstream tasks. It's worth noting that all previous efforts in hallucination detection have exclusively relied on manual annotation to address this challenge. 
%Finally, we introduce $\mathcal{FACTOID}$ (\ul{FAC}tual en\ul{T}ailment f\ul{O}r halluc\ul{I}nation \ul{D}etection), a benchmark for factual entailment techniques to detect factual incorrectness and highlight textual segments responsible for such errors, thus providing a comprehensive tool for assessing and improving the factual integrity of LLM-generated content.

\end{abstract}
% $\mathcal{ACORN}$ (\ul{A}utomatic Hallu\ul{C}inati\ul{O}n Evaluation for La\ul{R}ge La\ul{N}guage Models)

%Despite the advancements in Large Language Models, they remain susceptible to the challenge of hallucination. Numerous studies aim to gain a deeper understanding of this issue by initially identifying and subsequently addressing it. Various benchmarks have been suggested for conducting diverse experiments. Textual entailment is a prevalent approach for evaluating these benchmarks. Nevertheless, there is currently no automated metric available for quantifying hallucination. Hence, in this paper, we introduce $\mathcal{ACORN}$ (\ul{H}allucination \ul{A}utomat\ul{I}c Eva\ul{L}uation), a pioneering automatic entailment technique that examines the textual \textit{span} to address this gap.

\vspace{-10mm}
\begin{defin}

\begin{itemize}
[labelindent=-0.6em,labelsep=0.1cm,leftmargin=*]
\setlength\itemsep{0em}
\begin{spacing}{0.5}
\item[$\blacktriangleright$] 
{\footnotesize 
{\fontfamily{phv}\fontsize{8}{9}
%\begin{spacing}{1}
\selectfont
Introducing a new type of TE called "\ul{Factual Entailment (FE)}", aims to detect factual inaccuracies in content generated by LLMs while also highlighting the specific text segment that contradicts reality. (cf. \cref{sec:intro}).}
}

\item[$\blacktriangleright$] 
{\footnotesize 
{\fontfamily{phv}\fontsize{8}{9}\selectfont
Presenting $\mathcal{FACTOID}$ (FACTual enTAILment for hallucInation Detection), a benchmark dataset for FE (cf. \cref{sec:factoid}).}
}

\item[$\blacktriangleright$] 
{\footnotesize 
{\fontfamily{phv}\fontsize{8}{9}\selectfont
We propose an MTL framework for FE, yielding 30\% improvement in accuracy on the $\mathcal{FACTOID}$ benchmark compared to SoTA TE methods (cf. \cref{sec:fe}).}
}

\item[$\blacktriangleright$] 
{\footnotesize 
{\fontfamily{phv}\fontsize{8}{9}\selectfont
We assessed 15 modern LLMs and ranked them using our proposed \emph{\ul{Auto Hallucination Vulnerability Index ($HVI_{auto}$)}} (cf. \cref{sec:auto-hvi}).}
}

\vspace{-6mm}
\end{spacing}
\end{itemize}

\end{defin}

\vspace{-6mm}
\section{FACTUAL Entailment: The Necessity} \label{sec:intro}
\begin{comment}
\begin{figure*}
    \centering
    \includegraphics[width=\textwidth]{img/fac.pdf}
    \caption{Traditional Textual Entailment vs. Factual Entailment. In part A (top), we highlight the limitation of the TE method to identify as a refute case. However, in part (B), \acb{$\mathcal{ACORN}$ follows \acb{a multitask learning} approach by characterizing span-based nuanced factual entailment. It seeks to generate both an entailment score along with the span of the entailment.} Hence, this new entailment method helps in detecting the hallucination.}
    \label{fig:fac}
\end{figure*}
\end{comment}

Large generative AI models like GPT \citep{brown2020language,openai2023gpt4}, Stable Diffusion \cite{rombach2022high}, DALL-E \cite{ramesh2021zero,ramesh2022hierarchical}, and Midjourney \cite{midjourney}, face various challenges related to the risk of potential misuse. One such major challenge of Large Language Models (LLMs) is generating factually incorrect responses, which is referred to as \textit{hallucination}. 
%Several detection techniques proposed in \cite{manakul2023selfcheckgpt,mündler2023selfcontradictory} have attempted to detect hallucinations in LLMs. 
Recently, numerous techniques for mitigating hallucinations have been proposed, including i) Retrieval Augmented Generation \cite{peng2023check,vu2023freshllms,kang2023ever,gao2023rarr}, ii) Self Refinement through Feedback and Reasoning \cite{si2022prompting,mundler2023self,chen2023dress}, iii) Prompt Tuning \cite{cheng2023uprise,jones2023teaching}, iv) Introducing a New Decoding Strategy \cite{chuang2023dola,li2023inference}, v) Utilization of Knowledge Graph \cite{bayat2023fleek}, vi) Introducing Faithfulness based Loss Function \cite{yoon-etal-2022-information,qiu2023detecting}, and vii) Supervised Finetuning \cite{elaraby2023halo,tian2023finetuning,qiu2023think}. 

%Similarly, there are also several works in the direction of mitigating them \cite{varshney2023stitch}. 

%\textbf{Hallucinations types} Hallucination can occur in various forms. The generation of fictional characters is addressed in \cite{ladhak2023pre}. Similarly, the creation of imaginary locations is discussed in \cite{ladhak2023pre}. Additionally, a comprehensive taxonomy covering \textit{six} hallucination categories is introduced in \cite{rawte2023troubling}.

%\textbf{RAG is emerging as a solution} LLMs face several notable challenges, including being static, lacking domain-specific knowledge, and being perceived as opaque ``black boxes'' that are not easily understood. 

Hallucination mitigation has received considerable research attention recently, with Retrieval Augmented Generation (RAG) being considered the most promising approach to eliminate hallucinations in LLM generation. The working principle of RAG involves providing a prompt $p_1$ to the LLM, which generates text $t_1$. Since the LLM's factual knowledge is limited to its training data, it retrieves relevant documents or information $(r_1, r_2, r_3, ..., r_n)$ from a repository or search engine. This retrieved information is then used as context when generating the text from the LLM. Recent research suggests that RAG can effectively mitigate hallucinations to a certain extent \cite{}. However, this area is still evolving, and we anticipate further progress soon. Nonetheless, we argue that before and after applying any mitigation technique, it's crucial to understand the hallucination rate. Automatic hallucination detection is essential in this regard.

A straightforward solution to this could be to utilize state-of-the-art textual entailment (TE) techniques and adapt them for hallucination detection. The three possible outcomes of any TE method are (i) support/entailment, (ii) contradict/refute, and, (iii) neutral/not enough information. However, we have empirically demonstrated that SoTA TE techniques have significant shortcomings in terms of detecting factual errors in LLM-generated text. While the lack of entailment could signal the occurrence of hallucination, it should not be misconstrued as a definitive indicator of whether hallucination exists. For instance, what if both the first and second sentences are hallucinated? In that case, the fact that the sentences are entailed does not convey actionable insight as to whether hallucination is present. Similarly, the lack of entailment does not automatically mean that hallucination is occurring; it may simply indicate that the information provided is insufficient or that the texts are discussing different aspects of a topic. Therefore, a more nuanced approach is needed. This approach requires a combination of textual entailment recognition, factual verification, and span detection to mark the specific sections of both source and target text that contradict each other. One such scenario has been illustrated in \cref{fig:fig1}.

\section{Types of Hallucination} \label{sec:types}
Recent studies \cite{NEURIPS2022_df438caa, maynez-etal-2020-faithfulness, ladhak2023pre, raunak-etal-2021-curious} have explored various types of hallucinations. Building upon the work of \cite{rawte2023troubling}, we adopted their comprehensive categorization of hallucination types. We further streamlined this taxonomy, discarding a few rare categories. The hallucination categories we consider are as follows:

\subparagraph{Bothersome Numbers (BN):} This occurs when an LLM generates fictional numerical values (such as price, age, date, etc.).

\vspace{-3mm}

\begin{tcolorbox}
[boxsep=0pt,left=5pt,right=5pt,top=5pt,bottom=5pt,colback=DarkCyan!5!white,colframe=DarkCyan!35!black]
\scriptsize
\textbf{Original:} Patrick Mahomes, the Kansas City quarterback, dazzled in his team's Super Bowl win over the Eagles...  

  \vspace{-2mm}
  \DrawLine

\textbf{\scriptsize AI-generated:} He completed 26-of-38 passes for \textcolor{red}{286 yards} and two touchdowns ...

\vspace{-2mm}
\DrawLine

\textbf{\scriptsize Fact:} ...he added the second Super Bowl victory of his career, throwing for 182 yards and...
\end{tcolorbox}

\vspace{-4mm}

\subparagraph{Temporal Issue (TI):} This problem involves LLMs generating text that combines events from different timelines.

\vspace{-3mm}

\begin{tcolorbox}
[boxsep=0pt,left=5pt,right=5pt,top=5pt,bottom=5pt,colback=DarkCyan!5!white,colframe=DarkCyan!35!black]
\scriptsize
\textbf{Original:} Jurgen Flimm, who led some of Europe 2019s most important theaters, died on Feb. 4

  \vspace{-2mm}
  \DrawLine

\textbf{\scriptsize AI-generated:} In 1991, \textcolor{red}{Jurgen Flimm} was appointed artistic director of the Salzburg Festival.

\vspace{-2mm}
\DrawLine

\textbf{\scriptsize Fact:} Gerard Mortier was appointed as Artistic Director on 1 September 1991.
\end{tcolorbox}

\vspace{-4mm}

\subparagraph{ Imaginary Figure (IF):} This happens when an LLM fabricates a fictional persona without any concrete evidence.

\vspace{-3mm}

\begin{tcolorbox}
[boxsep=0pt,left=5pt,right=5pt,top=5pt,bottom=5pt,colback=DarkCyan!5!white,colframe=DarkCyan!35!black]
\scriptsize
\textbf{Original:} Russia pounded the front line in Ukraine's east and south with deadly artillery strikes...

  \vspace{-2mm}
  \DrawLine

\textbf{\scriptsize AI-generated:} The shelling is intense and non-stop, said local resident \textcolor{red}{Yevgeny Kondratyuk} ...

\vspace{-2mm}
\DrawLine

\textbf{\scriptsize Fact:} Yevgeny Kondratyuk does not exist!
\end{tcolorbox}

\vspace{-4mm}

\subparagraph{Place (P):} This issue occurs when LLMs generate an incorrect location related to an event.

\vspace{-3mm}

\begin{tcolorbox}
[boxsep=0pt,left=5pt,right=5pt,top=5pt,bottom=5pt,colback=DarkCyan!5!white,colframe=DarkCyan!35!black]
\scriptsize
\textbf{Original:} ...Another powerful earthquake struck Turkey and Syria on Monday, January 24, 2023...

  \vspace{-2mm}
  \DrawLine

\textbf{\scriptsize AI-generated:} 8 quake struck at 1:41 pm local time (1041 GMT) near the \textcolor{red}{city of Elazig in eastern Turkey}...

\vspace{-2mm}
\DrawLine

\textbf{\scriptsize Fact:} The quake struck in Hatay, Turkey’s southernmost province, and was measured at 6.4 magnitude...
\end{tcolorbox}
% \adil{not a good example}

\vspace{-2mm}

In this instance, the expression \emph{giant leap for humanity} is quoted from Neil Armstrong's renowned historical statement upon stepping onto the moon. 
% The issue of referencing the source of hallucination has also been recently explored in the work of  \cite{agrawal2023language}.

%\vspace{-2mm}
\section{Choice of LLM} \label{sec:llm}

We have chosen 15 modern LLMs that consistently exhibit excellent performance across various NLP tasks, as per the Open LLM Leaderboard \cite{open-llm-leaderboard}. The list includes: (i) GPT 4 \cite{openai2023gpt4}, (ii) GPT 3.5 \cite{ChatGPT}, (iii) Falcon \cite{almazrouei2023falcon}, (iv) GPT 2 \cite{radford2019language}, (v) MPT \cite{wang2023multitask}, (vi) OPT \cite{zhang2022opt}, (vii) LLaMA \cite{touvron2023llama}, (viii) BLOOM \cite{scao2022bloom}, (ix) Alpaca \cite{alpaca}, (x) Vicuna \cite{vicuna2023}, (xi) Dolly \cite{dolly}, (xii) StableLM \cite{liu2023your}, (xiii) XLNet \cite{yang2019xlnet}, (xiv) T5 \cite{raffel2020exploring}, and (xv) T0 \cite{DBLP:conf/iclr/DeleuKFKBLB22}.

\section{$\mathcal{FACTOID}$: Factual Entailment Dataset}   \label{sec:factoid}
We present $\mathcal{FACTOID}$ (FACTual enTAILment for hallucInation Detection), a benchmark dataset for FE containing 
total containing 2 million text pairs. Details are given in \cref{tab:factoid_dataset}. $\mathcal{FACTOID}$ is a synthetic extension of HILT dataset introduced by \cite{rawte2023troubling}. HiLT comprises a total of 492K sentences, out of which 129K are annotated for hallucination, indicating that 364K sentences are factually correct. At this juncture, we aim to synthesize these 129K sentences further for the factual entailment (FE) task. To accomplish this, we devise hallucination category-specific techniques, as detailed below:

\begin{comment}
\begin{table}[!htbp]
\centering
\scriptsize
\resizebox{0.62\columnwidth}{!}{
    \begin{tabular}{cc}  \toprule
    \bf Hallucination Category  & \bf \# sentences   \\ \midrule
    \bf Person   &  9,570  \\ 
    \bf Location     &  32,190  \\ 
    \bf Number  &  11,745 	\\ 
    \bf Time  &  36,105 	\\  \midrule
    \bf Total  &  89,610  \\   \bottomrule
    \end{tabular}
    }
    \vspace{-3mm}
    \caption{Statistics of the dataset.}
    \label{tab:data-stats}
\end{table}
%\vspace{-5mm}
\end{comment}

%For training and assessing the factual entailment (FE) module, we've compiled a dataset primarily using the HILT dataset introduced by \cite{rawte2023troubling}. HILT consists of prompts sourced from the New York Times, AI-generated text from 15 different LLMs, and annotations delineating specific types of hallucination at the sentence level. Overall, the dataset comprises 129K annotated sentences, with a total of 492K sentences included, as provided by the authors.
%$\mathcal{FACTOID}$ is a first-of-its-kind publicly available hallucination dataset. To construct this dataset, we have utilized two primary sources of data as prompts: (i) NYTimes tweets \cite{nyt}.

%\subsection{Sythesizing \emph{refute} FE samples}

\vspace{-1.5mm}
\paragraph{Bothersome Numbers (BN):} \textls[-10]{The HiLT dataset contains 7275 sentences associated with number-related hallucinations. Our aim is to produce more negative samples for Factual Entailment (FE) by randomly adjusting numbers in these sentences. However, mere number changes might not consistently create valid entailment scenarios. To overcome this, we applied automatic paraphrasing techniques (explained in Section X). Numbers were detected using regular expressions and altered randomly within a range of $\pm20\%$, as shown by the blue-marked numbers in the example. These paraphrased sentences effectively refute the originals.}
%There are total 7275 sentences in HiLT dataset for number related hallucination. Picking any one these sentences and further randomly changing number to generate more negative samples for FE is our motive. However only changing a number maynot be perfect entailment problem, therefore we did it through automatic paraphrasing, see section x for details. Numbers are identified through regular expression and changed randomly within $\pm20\%$ range, see blue marked numbers in the example. All these newly generated paraphrased sentences are refuting the original sentence. 

\vspace{-2mm}
\begin{tcolorbox}
[boxsep=0pt,left=2.5pt,right=5pt,top=5pt,bottom=5pt,colback=orange!15!white,colframe=orange!45!black]
\scriptsize
\tframed[line width=0.5bp,fill=Navy!65]{\textcolor{white}{\textbf{Original sentence}}} \tframed[line width=0.5bp,fill=DarkCyan!100]{\textcolor{white}{\textbf{The layoffs come after Twitter announced earlier this}}} \\ 
\tframed[line width=0.5bp,fill=DarkCyan!100]{\textcolor{white} {\textbf{ month that it would be cutting its global workforce by 8\% of people.}}}

\vspace{-1mm}
  \DrawLine

%\tframed[line width=0.5bp,fill=Navy!40]{\textcolor{white}{\textbf{Para \S 1}}} The layoffs follow Twitter's recent announcement that it will reduce its global workforce by \tframed[fill=Cyan!50]{12\%}.

%\tframed[line width=0.5bp,fill=Navy!40]{\textcolor{white}{\textbf{Para \S 2}}} The layoffs occurred subsequent to Twitter's earlier announcement this month about reducing its global workforce by \tframed[fill=Cyan!50]{21\%}.

\tframed[line width=0.5bp,fill=Navy!40]{\textcolor{white}{\textbf{Para \S 1}}} The job cuts were implemented following Twitter's announcement earlier this month that it would reduce its global workforce by \tframed[fill=Cyan!50]{10\%}.

\tframed[line width=0.5bp,fill=Navy!40]{\textcolor{white}{\textbf{Para \S 2}}} The layoffs were initiated subsequent to Twitter's earlier declaration this month regarding its plan to reduce its global workforce by \tframed[fill=Cyan!50]{4\%}.

\tframed[line width=0.5bp,fill=Navy!40]{\textcolor{white}{\textbf{Para \S 3}}} The staff reductions occurred subsequent to Twitter's earlier announcement this month about trimming its global workforce by \tframed[fill=Cyan!50]{2\%}.

\end{tcolorbox}

%\vspace{-20mm}
\paragraph{Temporal Issue (TI):} %The HiLT dataset consists of approximately 7,500 sentences, including the Time Wrap category from Factual Mirage, related to time-related hallucinations. Our objective is to create more negative samples for Factual Entailment (FE) by randomly changing the entities of two individuals from distinct time periods within these sentences. Recent studies suggest that LLMs acquire linear representations of space and time across various scales \cite{gurnee2023language}. We drew inspiration from previous research to design our experiment. The overall setup of the experiment is semi-automatic, meaning it requires human intervention.
The HiLT dataset, containing about 7,500 sentences from the Time Wrap category of Factual Mirage, focuses on time-related hallucinations. Our goal is to expand negative samples for FE by randomly altering the entities of two individuals from different time periods within these sentences. Recent studies indicate that LLMs grasp linear representations of space and time across various scales \cite{gurnee2023language}, which inspired our experiment design. The experiment setup is semi-automatic, requiring human intervention.

\textls[-10]{We identified an entity and manually formulated a question: ``\emph{When did the Amber Alert program start?}" We posed this question to an LLM and received the response: ``\emph{The Amber Alert program officially began in 1996.}" Subsequently, we randomly selected a number between 50 and 150 and subtracted it from 1996 to determine the desired timeframe, which in this case (\emph{let's assume}) was 1806. We then asked the LLM, ``\emph{Who was the President of the USA in 1806?}" and received the answer: ``\emph{Thomas Jefferson}." We substituted ``\emph{Obama}" with ``\emph{Jefferson}" in all automatically generated paraphrases. We chose Llama for this task based on its usage in prior research \cite{gurnee2023language}. Although this process required manual intervention, we were able to manage the generation process with two student annotators over a two-week period, given the 7.5K sentences in the dataset.}

\vspace{-2mm}
\begin{tcolorbox}
[boxsep=0pt,left=2.5pt,right=5pt,top=5pt,bottom=5pt,colback=orange!15!white,colframe=orange!45!black]
\scriptsize
\tframed[line width=0.5bp,fill=Navy!65]{\textcolor{white}{\textbf{Original sentence}}} \tframed[line width=0.5bp,fill=DarkCyan!100]{\textcolor{white}{\textbf{The Obama administration shut down the Amber Alert}}} \\ 
\tframed[line width=0.5bp,fill=DarkCyan!100]{\textcolor{white} {\textbf{ program because of the government shutdown in October 2013.}}}
%\tframed[line width=0.5bp,fill=DarkCyan!100]{\textcolor{white} {\textbf{}}}

\vspace{-1mm}
  \DrawLine

%\tframed[line width=0.5bp,fill=Navy!40]{\textcolor{white}{\textbf{Para \S 1}}} The Amber Alert program was discontinued during the \tframed[fill=Cyan!50]{Jefferson} administration due to the government shutdown in October 2013.

\tframed[line width=0.5bp,fill=Navy!40]{\textcolor{white}{\textbf{Para \S 1}}} Due to the government shutdown in October 2013, the \tframed[fill=Cyan!50]{Jefferson} administration ceased the operation of the Amber Alert program.

%\tframed[line width=0.5bp,fill=Navy!40]{\textcolor{white}{\textbf{Para \S 3}}} The Amber Alert program ceased operations during the government shutdown in October 2013 under the \tframed[fill=Cyan!50]{Jefferson} administration.

\tframed[line width=0.5bp,fill=Navy!40]{\textcolor{white}{\textbf{Para \S 2}}} During the government shutdown in October 2013, the \tframed[fill=Cyan!50]{Jefferson} administration made the decision to suspend operations of the Amber Alert program.

\tframed[line width=0.5bp,fill=Navy!40]{\textcolor{white}{\textbf{Para \S 3}}} During the government shutdown in October 2013, under the \tframed[fill=Cyan!50]{Jefferson} administration, the Amber Alert program halted.

\end{tcolorbox}

%We have applied automatic paraphrasing technique, see details in Section X. For a given prompt, we utilize Named Entity Recognition (NER) \cite{} to identify person names. Next, we employ a pre-trained Word2Vec-based euclidian distance measure to identify other person names that are close in vector space. We define an experimental Euclidean threshold for this analysis.

\vspace{-6mm}
\paragraph{Imaginary Figure (IF):} The HiLT dataset contains 15K sentences focusing on person-related hallucinations, particularly from the Generated Golem category in Factual Mirage. Our aim is to enhance negative samples for Factual Entailment (FE) by randomly altering the names of individuals in these sentences. We utilize an automatic paraphrasing technique detailed in Section X for this task. Named Entity Recognition (NER) \cite{} helps us identify person names within prompts. Then, leveraging a pre-trained Word2Vec-based \cite{mikolov2013distributed} Euclidean distance measure, we locate other person names in close vector space proximity. An experimental Euclidean threshold guides this process.

\vspace{-2mm}
\begin{tcolorbox}
[boxsep=0pt,left=2.5pt,right=5pt,top=5pt,bottom=5pt,colback=orange!15!white,colframe=orange!45!black]
\scriptsize
\tframed[line width=0.5bp,fill=Navy!65]{\textcolor{white}{\textbf{Original sentence}}} \tframed[line width=0.5bp,fill=DarkCyan!100]{\textcolor{white}{\textbf{One rescuer, \tframed[fill=red!100]{Hasan Cetin}, said he was motivated by thr}}} \\ 
\tframed[line width=0.5bp,fill=DarkCyan!100]{\textcolor{white} {\textbf{thought of the survivors he helped save.}}}

\vspace{-1mm}
  \DrawLine

%\tframed[line width=0.5bp,fill=Navy!40]{\textcolor{white}{\textbf{Para \S 1}}} One responder, \tframed[fill=Cyan!100]{Asif Iqbal}, mentioned that he was driven by the idea of the survivors he assisted in rescuing.

%\tframed[line width=0.5bp,fill=Navy!40]{\textcolor{white}{\textbf{Para \S 2}}} \tframed[fill=Cyan!100]{Mohammed Yunis}, a rescuer, expressed that the well-being of the survivors he aided was a motivating factor for him.

\tframed[line width=0.5bp,fill=Navy!40]{\textcolor{white}{\textbf{Para \S 1}}} \tframed[fill=Cyan!100]{Kader Hairat}, a courageous rescuer, shared his heartfelt sentiments regarding his noble actions.

\tframed[line width=0.5bp,fill=Navy!40]{\textcolor{white}{\textbf{Para \S 2}}} \tframed[fill=Cyan!100]{Safiq Masin} expressed that the primary driving force behind his heroic endeavors was the well-being of the survivors %he 
%selflessly aided.

\tframed[line width=0.5bp,fill=Navy!40]{\textcolor{white}{\textbf{Para \S 3}}} With compassion and determination, \tframed[fill=Cyan!100]{Shifaq Zaman} tirelessly worked to ensure the safety and comfort of those in need, drawing inspiration from their resilience and strength in the face %of adversity.

\end{tcolorbox}

\paragraph{Place (P):} The HiLT dataset includes approximately 13K sentences related to location-related hallucinations, specifically from the Geographic Erratum category of the Factual Mirage dataset. Our objective is to create additional negative samples for Factual Entailment (FE) by randomly modifying the names of individuals mentioned in these sentences. We utilize similar techniques as those used for person names. Initially, we apply Named Entity Recognition (NER) \cite{} to identify location names within a given prompt. Subsequently, we utilize a pre-trained Word2Vec-based Euclidean distance measure to identify other location names that are distant in vector space. For this analysis, we establish an experimental Euclidean threshold.

\vspace{-2mm}
\begin{tcolorbox}
[boxsep=0pt,left=2.5pt,right=5pt,top=5pt,bottom=5pt,colback=orange!15!white,colframe=orange!45!black]
\scriptsize
\tframed[line width=0.5bp,fill=Navy!65]{\textcolor{white}{\textbf{Original sentence}}} \tframed[line width=0.5bp,fill=DarkCyan!100]{\textcolor{white}{\textbf{Five people were killed, including a patient and a}}} \\ 
\tframed[line width=0.5bp,fill=DarkCyan!100]{\textcolor{white} {\textbf{family member, after a medical airplane crashed in Nevada on Friday night,}}}
\\ 
\tframed[line width=0.5bp,fill=DarkCyan!100]{\textcolor{white} {\textbf{the company Care Flight said.}}}

\vspace{-1mm}
  \DrawLine

\tframed[line width=0.5bp,fill=Navy!40]{\textcolor{white}{\textbf{Para \S 1}}} Five individuals, including a patient and a family member, lost their lives in a medical airplane crash in \tframed[fill=Cyan!100]{Tokyo} on Friday night, as reported by Care Flight.

%\tframed[line width=0.5bp,fill=Navy!40]{\textcolor{white}{\textbf{Para \S 2}}} Five individuals, including a patient and a family member, lost their lives in a medical plane crash in \tframed[fill=Cyan!100]{Paris} on Friday night, according to Care Flight, the company involved.

%\tframed[line width=0.5bp,fill=Navy!40]{\textcolor{white}{\textbf{Para \S 3}}} Five fatalities, including a patient and a family member, resulted from a medical airplane crash in \tframed[fill=Cyan!100]{Mumbai} on Friday night, as reported by Care Flight.

\tframed[line width=0.5bp,fill=Navy!40]{\textcolor{white}{\textbf{Para \S 2}}} According to a statement by Care Flight, a medical aircraft crash in \tframed[fill=Cyan!100]{Oslo} on Friday night resulted in the deaths of five individuals, among them a patient and a family member.

\tframed[line width=0.5bp,fill=Navy!40]{\textcolor{white}{\textbf{Para \S 3}}} Care Flight, the company responsible for emergency medical services, reported that a total of five individuals tragically lost their lives in a plane crash in \tframed[fill=Cyan!100]{Melbourne} on Friday night. Among the victims were a patient who was being transported and a family member accompanying them.

\end{tcolorbox}

\vspace{-2mm}
\paragraph{Span marks:} \textls[-10]{During the synthetic data expansion process, we retained all replacement markers and marked the original sentences where certain entities were replaced. \emph{\ul{It's crucial to note that FE exclusively provides span output for the refute case. Additionally, in instances where no other person name is available in the retrieved documents for the IF scenario, FE marks only the original sentence.}}}

\textbf{Hallucination classes}: Given that $\mathcal{FACTOID}$ extends the HILT dataset, and since HILT already contains manually annotated categories, we simply transferred those categories directly to $\mathcal{FACTOID}$.
\vspace{-2mm}

\begin{figure*}[!ht]
        \centering
        \begin{subfigure}[b]{0.33\textwidth}
            \begin{tcolorbox}[boxsep=0pt,left=2.5pt,right=5pt,top=5pt,bottom=5pt,colback=Teal!10!white,colframe=Teal!45!black]
            \tiny
            \centering
                \textbf{sent1}: The sun sets behind the mountains, casting a warm glow across the landscape. The sky transforms into a canvas of vibrant hues, from fiery oranges to soft purples. The air becomes cooler as twilight descends upon the earth. Nature's evening symphony begins, with the chirping of crickets and the rustle of leaves in the gentle breeze. As night falls, the world settles into a peaceful slumber, awaiting the dawn of a new day.
            \end{tcolorbox}    
            \end{subfigure}
            \hfill
            \begin{subfigure}[b]{0.63\textwidth}
            \begin{tcolorbox}[boxsep=0pt,left=2.5pt,right=5pt,top=5pt,bottom=5pt,colback=Teal!10!white,colframe=Teal!45!black]
            \tiny
            \centering
                \textbf{sent5}: Behind the rugged peaks, the sun gracefully retreats, suffusing the landscape with a radiant warmth that caresses every contour of the earth. Across the vast expanse, the heavens burst into an array of vibrant colors, from the fiery embrace of oranges to the tranquil embrace of purples, painting a captivating tableau above. As daylight wanes, a gentle chill creeps into the air, heralding the arrival of twilight, a transitional phase where the world pauses to catch its breath. Nature, in its evening chorus, serenades the fading light with the rhythmic chirping of crickets and the soft whispers of leaves dancing in the breeze. And so, with the advent of night, the world succumbs to a tranquil slumber, embracing the promise of renewal with each passing moment until the dawn of a new day breaks upon the horizon.

            \end{tcolorbox}    
            \end{subfigure}
            \hfill
            \vskip\baselineskip
        \vspace{-4mm}
        \begin{subfigure}[b]{0.45\textwidth}
            \centering
            \includegraphics[width=\textwidth, height=5.5cm]{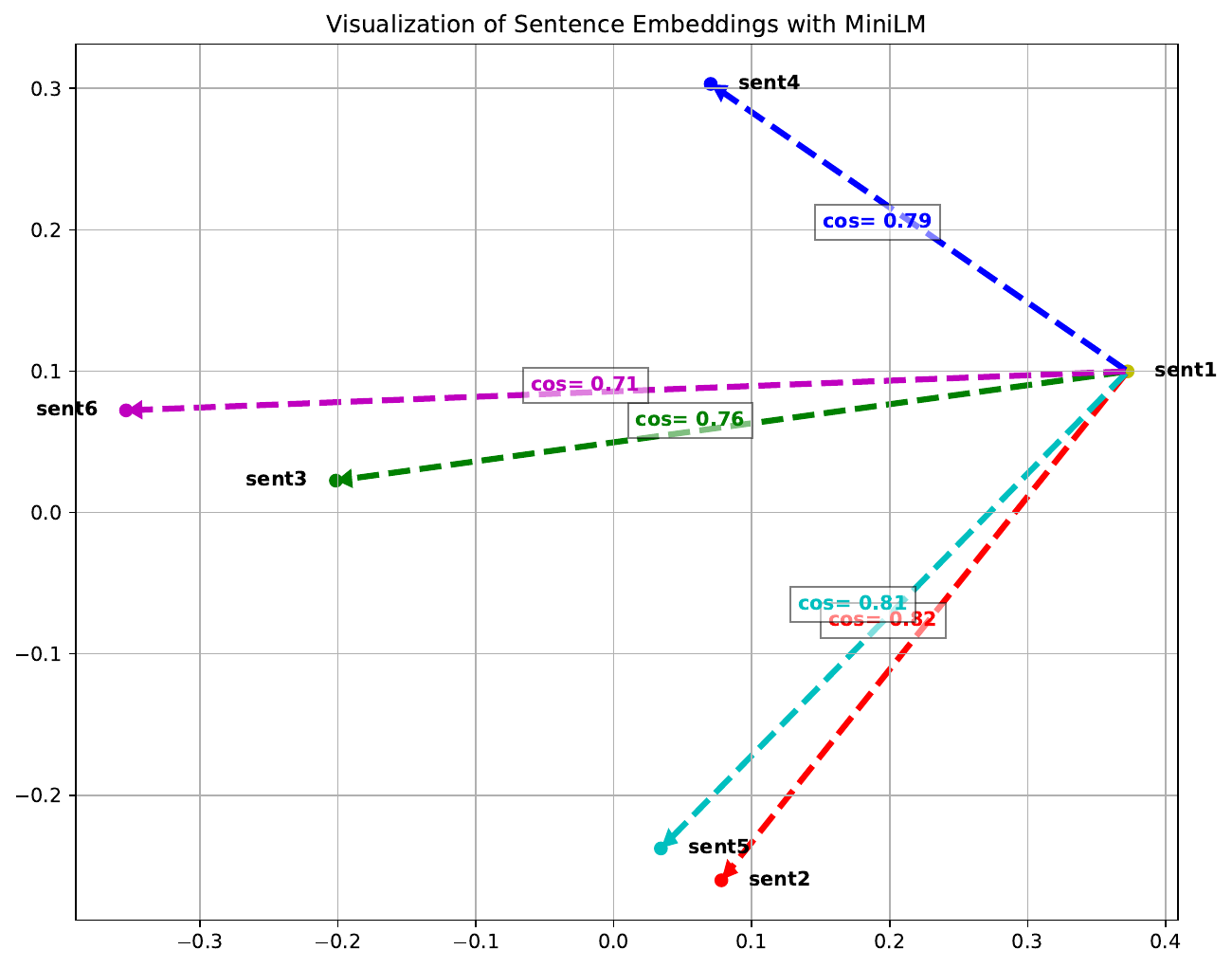}
            \caption[]%
            {{\small \textcolor{black}{Vanilla} sentence embedding.}}    
            \label{fig:net14}
        \end{subfigure}
        \hfill
        \begin{subfigure}[b]{0.45\textwidth}  
            \centering 
            \includegraphics[width=\textwidth, height=5.5cm]{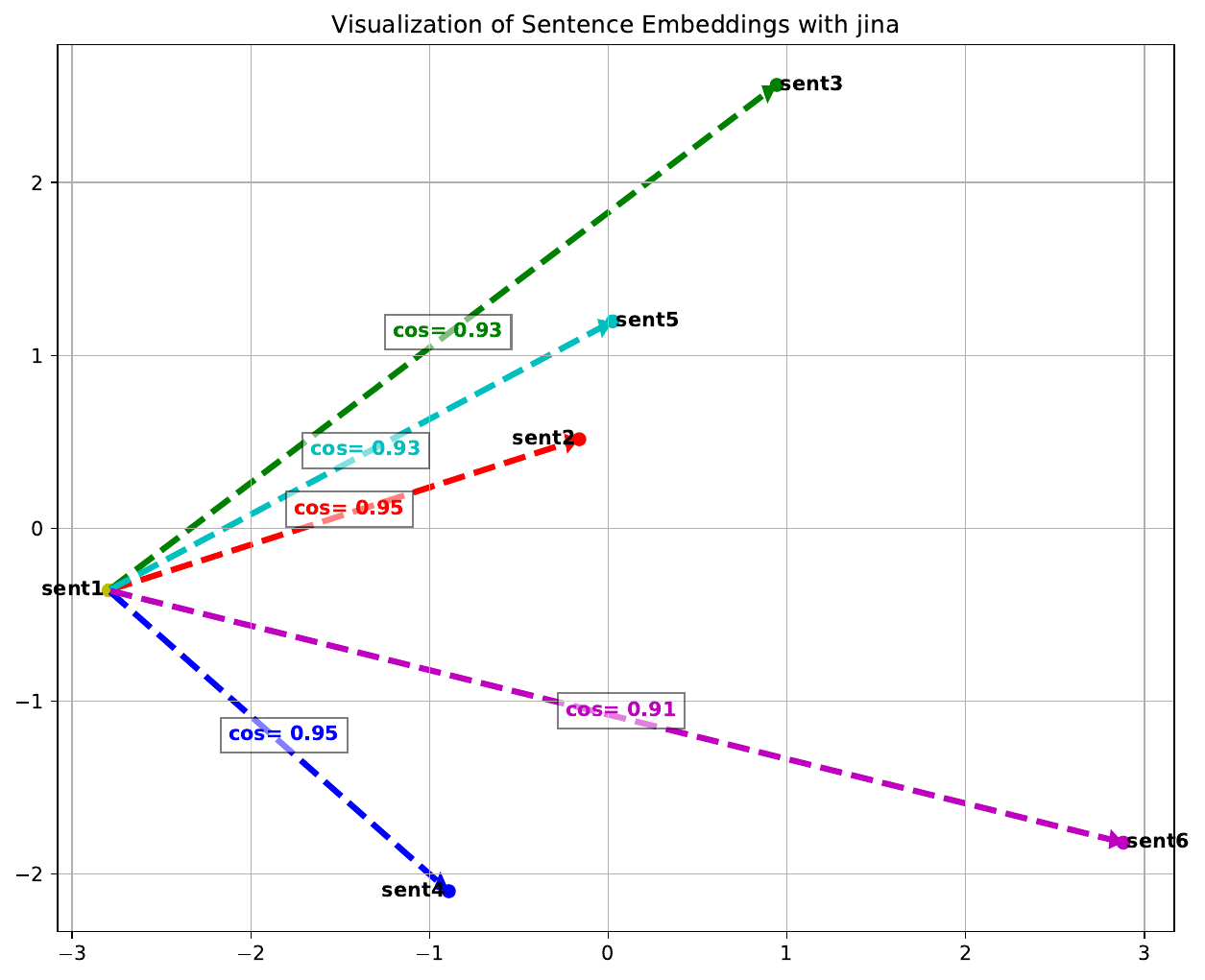}
            \caption[]%
            {{\small Longer sentence embedding.}}    
            \label{fig:net24}
        \end{subfigure}
        \caption[]%
            {{\small Utilizing longer embeddings for extended sentences is advantageous. The cosine similarities are more prominent in \texttt{Jina} embeddings \cite{günther2023jina} compared to \texttt{MiniLLM} \cite{gu2023knowledge}. Consequently, the cosine similarity for the pair \textbf{(sent1, sent2)} increases from 0.76 to 0.93, as indicated by the \textcolor{Green}{green} dashed line.}} 
        \label{fig:pause-ig}
\vspace{-2mm}        
\end{figure*}

\begin{figure*}
    \centering
    \includegraphics[width=\textwidth]{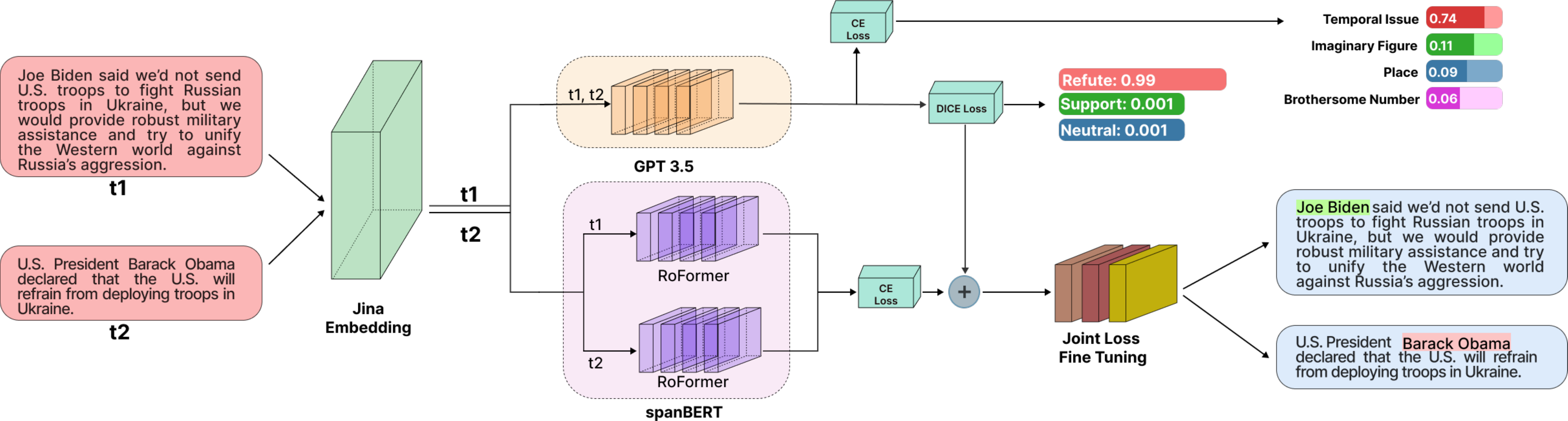}
    \vspace{-5mm}
    \caption{A summary of the overall multi-task learning framework for Factual Entailment. The framework encompasses three tasks: i) entailment, ii) span detection, and iii) hallucination classification.}
    \label{fig:mtl}
    \vspace{-2.5mm}
\end{figure*}

\subsection{Automatic Paraphrasing}
When choosing automatic paraphrasing, there are many other factors to consider for e.g., a model may only be able to generate a limited number of paraphrase variations compared to others, but others can be more correct and/or consistent. As such, we consider three major dimensions in our evaluation: \textit{(i) \textbf{Coverage}: a number of considerable generations, (ii) \textbf{ Correctness}: correctness in those generations, and (iii) \textbf{Diversity}: linguistic diversity in those generations}. We conducted experiments with three available models: (a) Pegasus \cite{zhang2020pegasus}, (b) T5 (T5-Large) \cite{raffel2020exploring}, and (c) GPT-3 (\texttt{text-davinci-003} variant) \cite{brown2020language}. Based on empirical observations, we concluded that GPT-3 outperformed all the other models. To offer transparency around our experiment process, we detail the aforementioned evaluation dimensions as follows.

\vspace{-2mm}
\begin{table}[H]
\centering
\resizebox{0.7\columnwidth}{!}{%
\begin{tabular}{@{}lcccc@{}}
\toprule
\textbf{Model}           &  \textbf{Coverage}  & \textbf{Correctness} & \textbf{Diversity} \\ \midrule
\textbf{Pegasus}          &   32.46   &    94.38\%       &     3.76      \\
\textbf{T5}               &  30.26       &      83.84\%       &    3.17       \\
\textbf{GPT-3} &   35.51   &   88.16\%      &     7.72      \\
\bottomrule
\end{tabular}%
}
\caption{Experimental results of automatic paraphrasing models based on three factors: \textit{(i) coverage, (ii) correctness, and (iii) diversity}; GPT-3 (\texttt{text-davinci-003}) is the most performant considering all three aspects.}
\label{tab:my-table}
\end{table}
\vspace{-4mm}

A comprehensive discussion regarding Coverage, Correctness, and Diversity, along with the experimental setup for paraphrasing, is available in \cref{sec:para}.

%\textcolor{red}{How the paraphrase setup has been utilized to generate sentence variation has been reported in the following algorithm and illustrated in the ~\ref{fig:acorn-para}.}
%\adil{we need a paragraph here}

\subsection{$\mathcal{FACTOID}$: Statistics}
$\mathcal{FACTOID}$ extends the HiLT dataset synthetically. HiLT encompasses a total of 492K sentences, with 129K annotated for hallucination, leaving 364K sentences deemed factually correct. As we exclusively expand the hallucinated sentences through paraphrasing, the resulting $\mathcal{FACTOID}$ dataset may suffer from class imbalance. To address this, we also expanded the 364K factually correct sentences. A statistical overview of $\mathcal{FACTOID}$ is presented in Table ~\ref{tab:factoid_dataset}.

% \usepackage{booktabs}
% \usepackage{graphicx}
\begin{table}[h!]
\centering
\resizebox{\columnwidth}{!}{%
\begin{tabular}{@{}lrrrr@{}}
\toprule
 &
  \multicolumn{1}{c}{\textbf{HILT}} &
  \multicolumn{1}{c}{\cellcolor[HTML]{FFFFFF}\textbf{Synthesized}} &
  \multicolumn{1}{c}{\cellcolor[HTML]{FFFFFF}\textbf{HILT}} &
  \multicolumn{1}{c}{\cellcolor[HTML]{FFFFFF}\textbf{Synthesized}} \\ \midrule
\textbf{Hallucination Type} &
  \multicolumn{2}{c}{\textbf{\# Positive Pairs}} &
  \multicolumn{2}{c}{\cellcolor[HTML]{FFFFFF}\textbf{\# Negative Pairs}} \\
\textbf{Imaginary Figure}  & 120800       & 507360       & 14800        & 62160       \\
\textbf{Place}             & 116770       & 513788       & 13050        & 56115       \\
\textbf{Bothersome Number} & 68570        & 281137       & 7275         & 40740       \\
\textbf{Temporal Issue}    & 57860        & 271942       & 6600         & 29700       \\
\textbf{Total}             & \multicolumn{2}{c}{1938227} & \multicolumn{2}{c}{230440} \\ \bottomrule
\end{tabular}%
}
\caption{$\mathcal{FACTOID}$ dataset statistics.}
\label{tab:factoid_dataset}
\vspace{-2mm}
\end{table}

\begin{comment}
\begin{figure*}[!]
\centering
\minipage{\textwidth}
% \minipage{\textwidth}
% \begin{table*}[!ht]%[H]
\vspace{-1mm}
\resizebox{0.98\textwidth}{!}{%
\begin{tabular}{p{0.2\textwidth}|p{1.5\textwidth}} \toprule
{ \textbf{Prompt}} 
& \textit{USA's stance on Ukraine Russia war} \\ \hline
{ \textbf{Hallucinated text}}
& 
  ...U.S. President \colorbox{pink}{Barack Obama} says the U.S. will not put troops in Ukraine... \\ \hline
{ \textbf{Fact}}
& 
 \colorbox{Khaki}{Joe Biden} said we’d not send U.S. troops to fight Russian troops in Ukraine, but we would provide robust military assistance and try to unify the Western world against Russia’s aggression. \\  \bottomrule
\end{tabular}%
}
\vspace{-2mm}
\caption{A hallucination example: \colorbox{pink}{Hallucinated span} vs. \colorbox{Khaki}{Factual span}.} 
\label{tab:hallucination_ex}
\endminipage
\end{figure*}
\end{comment}
\section{Factual Entailment - MTL approach} \label{sec:fe}
Multi-task learning is a widely-used approach in NLP to create end-to-end architectures that achieve multiple objectives simultaneously. In our work, we introduce several key contributions in terms of design choices, including the use of different LLMs for different tasks, employing long-text embedding, SpanBERT, RoFormer, and implementing specific loss functions as per the requirements of each task. Further details about these nuances are discussed below.

%As shown in \cref{fig:mtl}, Multi-task learning. Dateless merging like the deception paper and will merge two LLMs.
\vspace{-1mm}
\subsection{\textls[-10]{Long-Text High-Dimensional Embeddings}}

\textls[-10]{Long-text embeddings in NLP signify a transformative shift from traditional shorter embeddings, overcoming limitations and expanding application possibilities. Ranging from 768 to 4096 dimensions, these embeddings excel at capturing the semantics of extensive texts, enhancing document-level comprehension. They mitigate information loss by processing entire texts without truncation, preserving crucial context and details. Notably adept at grasping long-distance relationships, they prove invaluable for tasks like question answering and textual entailment, enabling sophisticated analyses in thematic development, stylistic evolution, and sentiment tracking. This advancement in NLP unlocks new potentials, offering a deep understanding for tasks requiring both holistic context comprehension and nuanced topical insight. Since entailment is a classification task, we chose \texttt{e5-mistral-7b-instruct} based on its top classification performance reported on the MTEB Leaderboard \cite{muennighoff2022mteb}. \cref{fig:pause-ig} illustrates the merits of using long-text embeddings for extended sentences compared to vanilla sentence embeddings.} Table \ref{tab:long-emb} offers a summary of long-text embedding models that were considered based on their classification performance on the MTEB Leaderboard:

\begin{table}[!ht]
\vspace{-1mm}
\centering
\resizebox{0.4\textwidth}{!}{%
\begin{tabular}{lll} \toprule
\textbf{Model} & \textbf{Length} &  \\ \midrule
\texttt{SFR-Embedding-Mistral} & 4096-dimensional embeddings over 32K tokens &  \\
\texttt{e5-mistral-7b-instruct} & 4096-dimensional embeddings over 32K tokens &  \\
\texttt{nomic-embed-text-v1} & 768-dimensional embeddings over 8K tokens &  \\
\texttt{text-embedding-3-large} & 3072-dimensional embeddings over 8K tokens & \\
\texttt{jina-embeddings-v2-base} & 8192-dimensional embeddings over 8K tokens & \\\bottomrule
\end{tabular}%
}
\vspace{-1mm}
\caption{Examples of long-text embedding models.}
\label{tab:long-emb}
\end{table}

% \adil{what experimental support we have to decide which embedding is performing best for our case?}
\vspace{-1mm}
\subsection{\textls[0]{Introducing Span-based Textual Entailment}}

%\acb{The example in \cref{tab:hallucination_ex} discusses a specific example where an LLM-generated text, in the context of the Russia-Ukraine war, mistakenly identifies \textit{Barack Obama} as the US President, while in reality, \textit{Joe Biden} was in office during this period. As indicated earlier in \cref{sec:intro}, The passage highlights that despite the text being scored as 'supportive' in terms of textual entailment, it still contains a factual inaccuracy or 'hallucination.'} 

%\acb{To address this, the passage suggests a refined approach to text analysis. Instead of evaluating entire sentences for textual entailment, focusing on specific spans of text could more accurately identify these inaccuracies. By doing so, one can distinguish between different entities (such as \textit{Barack Obama} and \textit{Joe Biden}) more effectively, leading to a clearer detection of hallucinations. This approach underlines the complexity of text analysis, especially in scenarios where nuanced understanding and contextual awareness are essential for accuracy.}
The example in \cref{fig:mtl} illustrates a case where an LLM, discussing the Russia-Ukraine war, incorrectly identifies \textit{Barack Obama} as the US President instead of \textit{Joe Biden}. Despite being deemed 'supportive' in textual entailment, the text contains a factual inaccuracy or `hallucination.' To improve accuracy, the passage suggests refining text analysis by focusing on specific spans rather than entire sentences. 
%This approach enhances the detection of inaccuracies involving different entities, emphasizing the intricate nature of text analysis, particularly in contexts requiring nuanced understanding and contextual awareness.

%\subsection{SpanBERT with RoFormer for Span Identification}
%\adil{this section need to be rewritten completely}

%One key challenge in the above approach is identifying the spans in a given text. Some possible approaches are using SpanBERT \cite{joshi-etal-2020-spanbert} or positional encoding. 
\paragraph{SpanBERT:} It is specifically designed to understand and represent spans of text \cite{joshi-etal-2020-spanbert}, making it useful for tasks involving relationships between different segments of a document or passage. It also enhances the capabilities of BERT by considering the context of spans, enabling a more nuanced understanding of language structure and meaning. 
\paragraph{RoFormer:} %Positional encoding can help maintain the \emph{word order}. Although there are different kinds of positional encoding techniques, one commonly used technique today is the Rotary Position Embedding (RoPE) \cite{SU2024127063}. \acb{RoPE employs distinct rotatory matrices based on the absolute position of each token. It calculates scores between keys and queries using relative position information, contributing to exceptional performance and prolonged decay in recent LLMs such as PaLM \cite{chowdhery2022palm} and LLaMA \cite{touvron2023llama}.} Contrary to conventional position embeddings, which are constrained by a predetermined sequence length, RoPE has the flexibility to adapt and function effectively with sequences of varying lengths. \acil{fact check this last line since absolute encoding also uses sinusoidal for theoretically infinite length.}
Introduced in \cite{su2022roformer}, utilizes a rotation matrix to encode absolute position while incorporating explicit relative position dependencies in self-attention formulation. This approach, featured in RoFormer, imparts beneficial properties such as sequence length flexibility, diminishing inter-token dependency with increasing relative distances, and the ability to integrate relative position encoding into linear self-attention.

\subsection{Loss Functions}
%\adil{Here we have write why we use specific loss functions at those places}
% \adil{need to relook}
% In exploring sub-task loss functions, we tested various options, including cross-entropy, focal loss, dice loss, and distribution-balanced loss (DB). 
We employed cross-entropy loss for span detection and hallucination type identification, while dice loss \cite{Sudre_2017} proved to be the best fit for entailment. Due to the significant imbalance in the \texttt{support} class, we opted for dice loss, known for its effectiveness in handling imbalanced datasets.
% For detailed results and an in-depth discussion of different loss functions, refer to Appendix D.2.

%During our exploration for suitable sub-task loss functions, we experimented with several available options, including (i) cross-entropy loss, (ii) focal loss \cite{lin2017focal}, (iii) dice loss \cite{li2019dice}, and (iv) distribution-balanced loss (DB) \cite{huang-etal-2021-balancing}. After a thorough evaluation, we observed that distribution-balanced loss yielded the best performance for layer 1, the cross-entropy loss was most effective for layer 2, focal loss performed well for layer 3, and dice loss was the optimal choice for layer 4. For a comprehensive overview of the results and an in-depth discussion of different loss functions, please refer to the Appendix D.2.

%\paragraph{Loss function} \textcolor{red}{taken from deception paper as is} During our exploration for suitable sub-task loss functions, we experimented with several available options, including (i) cross-entropy loss, (ii) focal loss (Lin et al., 2017), (iii) dice loss (Li et al., 2019), and (iv) distribution-balanced loss (DB) (Huang et al., 2021a). After a thorough evaluation, we observed that distribution-balanced loss yielded the best performance for layer 1, cross-entropy loss was most effective for layer 2, focal loss performed well for layer 3, and dice loss was the optimal choice for layer 4. For a comprehensive overview of the results and an in-depth discussion of different loss functions, please refer to the Appendix D.2.
\section{Performance of FE}  \label{sec:performance}
Our empirical findings depicted in \cref{fig:fe} illustrate that the proposed Factual Entailment (FE) outperforms the state-of-the-art textual entailment (TE) methods. Some key takeaways are listed below:

\begin{figure}[!ht]
    \centering
    \includegraphics[width=0.45\textwidth]{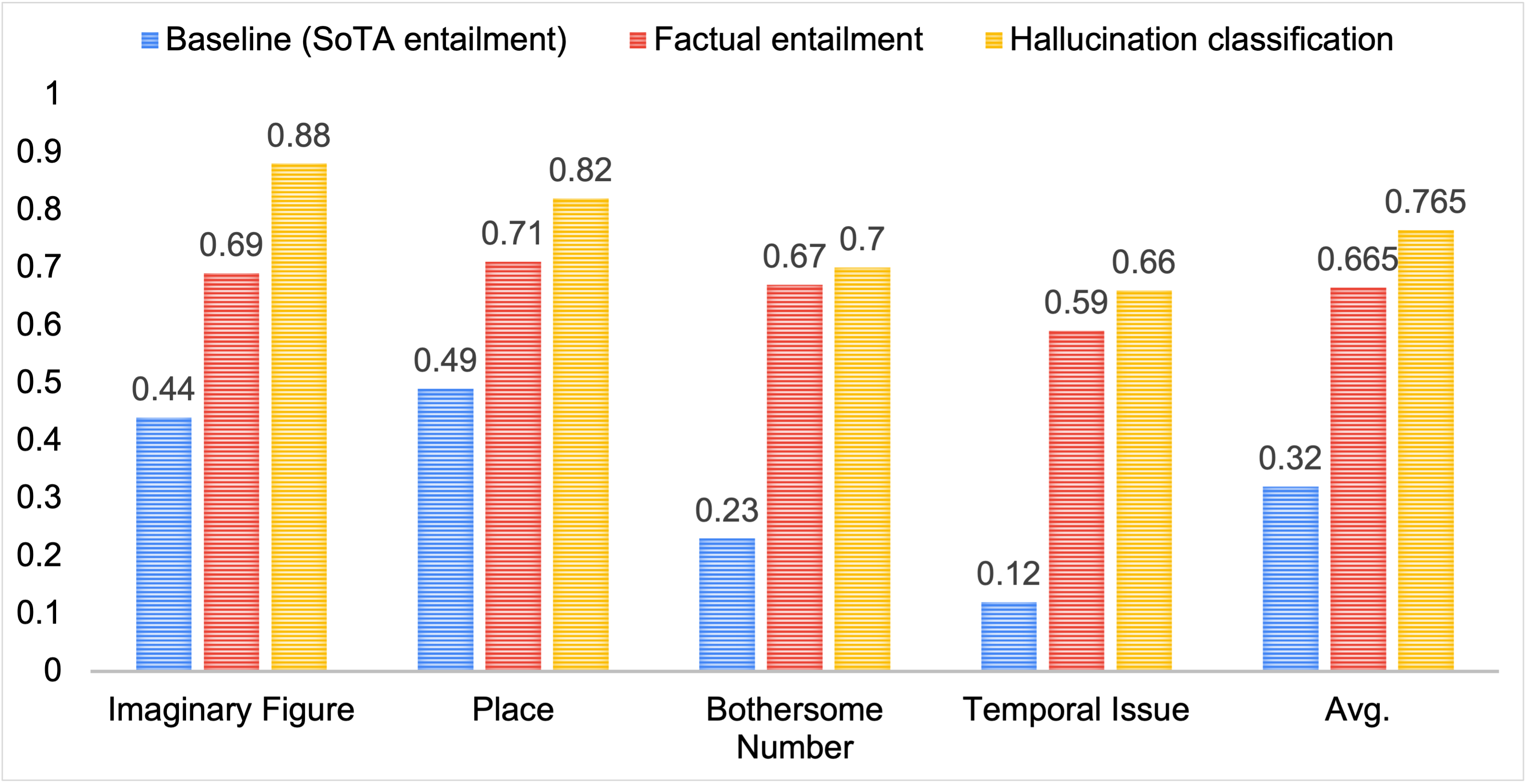}
    \caption{Results showing how FE performs better than TE at detecting hallucination in six different categories.}
    \label{fig:fe_results}
\vspace{-4mm}     
\end{figure}
\section{Automating Hallucination Vulnerability Index (HVI)}  \label{sec:auto-hvi}
\vspace{-1mm}
\textls[-10]{The Hallucination Vulnerability Index (HVI) was initially proposed by \cite{rawte2023troubling}. However, their approach relied entirely on manual annotation for HVI assessment. In this study, we introduce an automated hallucination metric, $HVI_{auto}$, as defined in \cref{eq:eq_1}. By automating the detection and classification of hallucinations, it is now feasible to calculate HVI automatically. To compute $HVI_{auto}$ for the LLMs discussed in Section \ref{sec:llm}, we leveraged 2,500 prompts from the HILT dataset \cite{rawte2023troubling}. These prompts were used to generate text from LLMs, and then Factual Entailment (FE) was applied to the generated text to detect hallucinations and classify them into different types.}
When defining $HVI_{auto}$, we take several factors into account. We consider $U$ as the total number of sentences we have in the corpus. Moreover, two/more LLMs can exhibit varying characteristics of hallucination, including person, location, time and number. For instance, if we have two LLMs and their total number of generated hallucinations in terms of sentences are the same, but $\text{LLM}_1$ produces significantly more time related hallucinations than $\text{LLM}_2$, we cannot rank them same. This comparative measure is achieved using multiplicative damping factors, $\delta_{BN}$, $\delta_{TI}$, $\delta_{IF}$ and $\delta_P$ which are calculated based on $\mu \pm rank_x \times \sigma$. Initially, we calculate the HVI for all the LLMs, considering $\delta_{BN}$, $\delta_{TI}$, $\delta_{IF}$ and $\delta_P$ as one. With these initial HVIs, we obtain the mean ($\mu$) and standard deviation ($\sigma$), allowing us to recalculate the HVIs for all the LLMs. The resulting HVIs are then ranked and scaled providing a comparative spectrum as presented in \cref{tab:hvi_spectrum}. Having damping factors enables easy exponential smoothing with a handful of data points, similar to z-score normalization \cite{Normalization-z} and min-max normalization \cite{Normalization-min}. Finally, for ease of interpretability, HVI is scaled between $0-100$.
\vspace{-0.5mm} 
\begin{equation}
\begin{split}
 \Scale[0.8]{
 HVI_{auto} = \frac{100}{U}[\sum_{x=1}^{U}(\delta_{BN}*H_{BN}+\delta_{TI}*H_{TI}}
 \\
 \Scale[0.8]{
 delta_{IF}*H_{IF}++\delta_{P}*H_{P}]
 }
\end{split}
\label{eq:eq_1}
\end{equation}

\vspace{-3mm}
\begin{figure}[h!]
\centering
\includegraphics[width=\columnwidth]{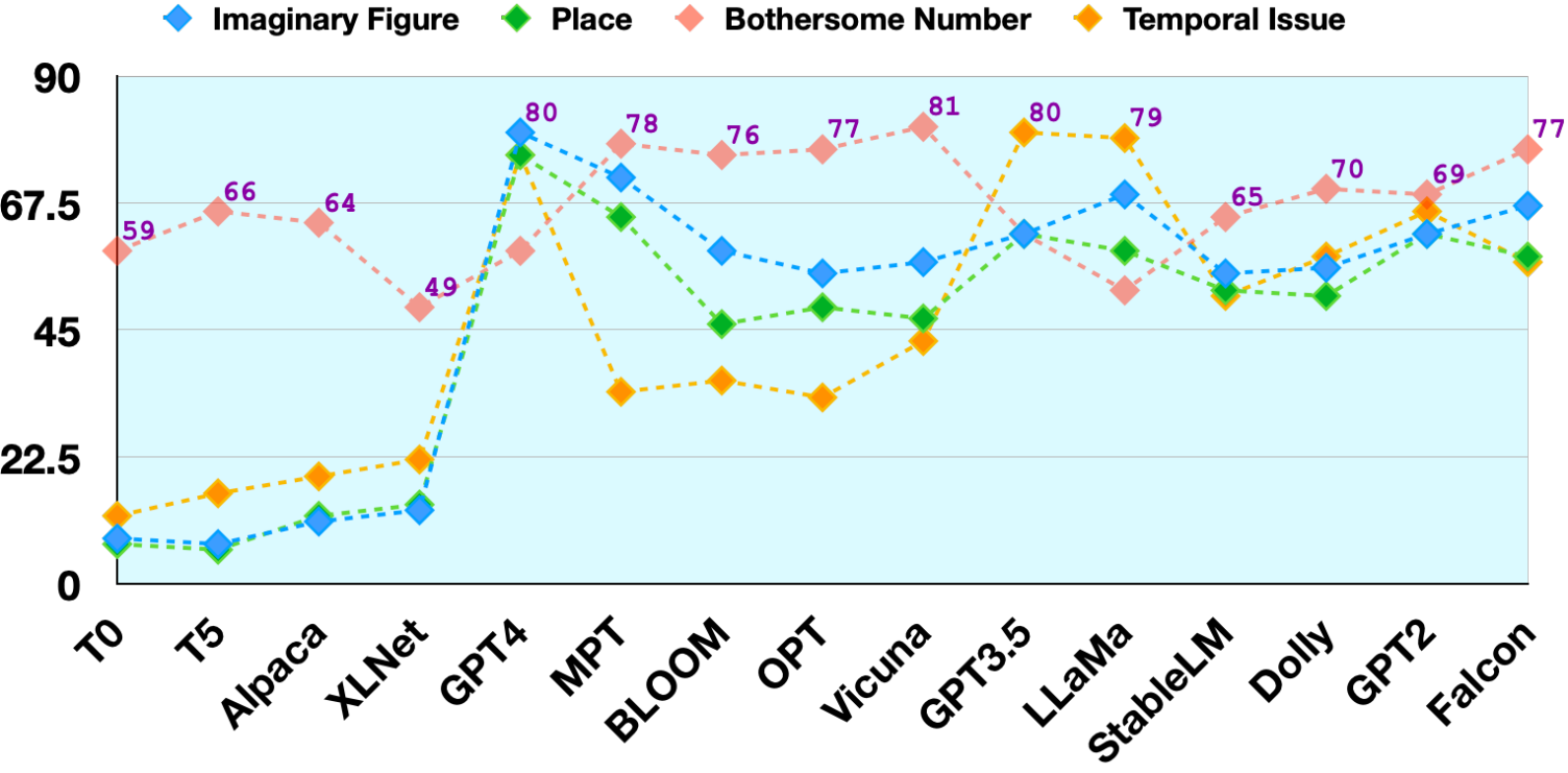}
\vspace{-5mm}
  \caption{HVI for different hallucination categories across various LLMs.}
  \label{fig:hvi-cat}
  %\vspace{-2mm}
\end{figure}

\vspace{-1mm}
\begin{figure}[h]
\minipage{0.4\textwidth}
\centering
%\begin{table}[!ht]
%\tiny
%\centering
%\resizebox{0.32\textheight}{!}{%
\resizebox{0.9\columnwidth}{!}{%
\centering
\begin{tabular}{lll}
\toprule
\textbf{LLM} & \textbf{Size} & \textbf{$\mathbf{HVI_{auto}}$ (0-100)}
\\ \midrule
\textbf{Falcon}    &  7B       &   80 - \begin{tikzpicture}
\centering
\draw[red, very thick, fill=red] (0,0)  rectangle (5.62,.1);
\end{tikzpicture}
\\
\textbf{GPT-2}    &   1.5B      &     78  - \begin{tikzpicture}
\centering
\draw[red, very thick, fill=red] (0,0)  rectangle (5.12,.1);
\end{tikzpicture}
\\
\textbf{Dolly}    &   12B    &    77  - \begin{tikzpicture}
\centering
\draw[red, very thick, fill=red] (0,0)  rectangle (4.38,.1);
\end{tikzpicture}
\\
\textbf{StableLM}    &  7B   &     73  - \begin{tikzpicture}
\centering
\draw[orange, very thick, fill=orange] (0,0)  rectangle (3.88,.1);
\end{tikzpicture}  
\\
\textbf{LLaMA}    &  65B    &     65 - \begin{tikzpicture}
\centering
\draw[orange, very thick, fill=orange] (0,0)  rectangle (3.69,.1);
\end{tikzpicture}   
\\
\textbf{GPT-3.5}   &    175B      &     65 - \begin{tikzpicture}
\centering
\draw[orange, very thick, fill=orange] (0,0)  rectangle (3.56,.1);
\end{tikzpicture}   
\\
\textbf{Vicuna}     & 13B      &   55   - \begin{tikzpicture}
\centering
\draw[orange, very thick, fill=orange] (0,0)  rectangle (3.31,.1);
\end{tikzpicture}
\\
\textbf{OPT}    &   175B   &     53 - \begin{tikzpicture}
\centering
\draw[violet, very thick, fill=violet] (0,0)  rectangle (3.06,.1);
\end{tikzpicture} 
\\
\textbf{BLOOM}   &   176B     &   52  - \begin{tikzpicture}
\centering
\draw[violet, very thick, fill=violet] (0,0)  rectangle (3,.1);
\end{tikzpicture}   
\\
\textbf{MPT}    &  7B    &     49 - \begin{tikzpicture}
\centering
\draw[violet, very thick, fill=violet] (0,0)  rectangle (2.94,.1);
\end{tikzpicture}  
\\
\textbf{GPT-4}    &   1.7T    &      46 - \begin{tikzpicture}
\centering
\draw[violet, very thick, fill=violet] (0,0)  rectangle (2.5,.1);
\end{tikzpicture}   
\\
\textbf{XLNet}   &   340M     &      45 - \begin{tikzpicture}
\centering
\draw[blue, very thick, fill=blue] (0,0)  rectangle (2.38,.1);
\end{tikzpicture}  
\\
\textbf{Alpaca}   &   65B    &     44 - \begin{tikzpicture}
\centering
\draw[blue, very thick, fill=blue] (0,0)  rectangle (2.25,.1);
\end{tikzpicture}  
\\
\textbf{T5}    &    11B      &     33  - \begin{tikzpicture}
\centering
\draw[blue, very thick, fill=blue] (0,0)  rectangle (2.25,.1);
\end{tikzpicture}  
\\
\textbf{T0}    &      11B    &     32 - \begin{tikzpicture}
\centering
\draw[blue, very thick, fill=blue] (0,0)  rectangle (2.0,.1);
\end{tikzpicture}  
\\
% ERNIE           &     40 - \begin{tikzpicture}
% \centering
% \draw[violet, very thick, fill=violet] (0,0)  rectangle (2,.5);
% \end{tikzpicture}  
% \\
% ALBERT          &   33 - \begin{tikzpicture}
% \centering
% \draw[blue, very thick, fill=blue] (0,0)  rectangle (1.8,.5);
% \end{tikzpicture}    
% \\
% DeBERTa         &     32 - \begin{tikzpicture}
% \centering
% \draw[blue, very thick, fill=blue] (0,0)  rectangle (1.6,.5);
% \end{tikzpicture}    
% \\
% ELECTRA         &    32 - \begin{tikzpicture}
% \centering
% \draw[blue, very thick, fill=blue] (0,0)  rectangle (1.8,.5);
% \end{tikzpicture}     
% \\
% RoBERTa         &    28 - \begin{tikzpicture}
% \centering
% \draw[blue, very thick, fill=blue] (0,0)  rectangle (1.4,.5);
% \end{tikzpicture}     
% \\
% BERT            &    26 - \begin{tikzpicture}
% \centering
% \draw[blue, very thick, fill=blue] (0,0)  rectangle (1,.5);
% \end{tikzpicture}     
% \ \ 
\bottomrule
\end{tabular}%
}
\label{ahvi}
\vspace{-1mm}
\caption{The HVI scale illustrates the hallucination tendencies exhibited by various LLMs.}
  \label{tab:hvi_spectrum}
\endminipage
\minipage{0.05\textwidth}
\vspace{-12mm}
\includegraphics[height=4.5cm]{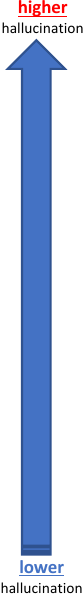}
\endminipage
\vspace{-4.5mm}
\end{figure}

\begin{tcolorbox}[
left=9pt,right=2pt,colback=Olive!5!white,colframe=Olive!75!black,colbacktitle=Olive,
  title=\footnotesize \fontfamily{qbk} \selectfont \textbf{Implications derived from $\mathbf{HVI_{auto}}$} ]
  
\vspace{-2mm}
\begin{itemize}
[labelindent=-0.2em,labelsep=0.1cm,leftmargin=*]
\setlength\itemsep{0em}
\begin{spacing}{0.85}

\item[\ding{118}] 
{\footnotesize 
{\fontfamily{phv}\fontsize{7}{8}
%\begin{spacing}{1}
\selectfont
Larger LLMs without RLHF \cite{DBLP:journals/corr/abs-1909-08593} are prone to hallucination, as shown in \cref{tab:hvi_spectrum}. 
}
%\end{spacing}
}
\vspace{-1mm} 
\item[\ding{118}] 
{\footnotesize 
{\fontfamily{phv}\fontsize{7}{8}
%\begin{spacing}{1}
\selectfont
Number-related issues are widespread across most LLMs, although they appear notably lower in certain models such as XLNet and StableLM. The reasons behind this discrepancy remain unclear and warrant further investigation in the future.}
%\end{spacing}
}
\vspace{-1mm} 
\item[\ding{118}] 
{\footnotesize 
{\fontfamily{phv}\fontsize{7}{8}
%\begin{spacing}{1}
\selectfont
Hallucination categories such as Imaginary Figures and Temporal issues tend to increase with the size of LLMs.
}
%\end{spacing}
}
%\end{spacing}

\vspace{-6mm}
\end{spacing}
\end{itemize}
\end{tcolorbox}
\vspace{-2mm}

\begin{comment}
\begin{equation} \nonumber
 \Scale[0.8]{HVI_x = \frac{100}{U*2}\left[ \sum_{x=1}^U(N(x)-N(E F M)) *\left(1-P(E F M)+\delta_1\right)+ \right.} \\
\Scale[0.8]{\left. (N(x)-N(E S L))*\left(1-P(E S L)+\delta_2\right)\right]}
\label{eq:eq1}
\end{equation}

The reranking of the SOTA LLMs is given below.

\vspace{-5mm}
\begin{figure}[h]
\minipage{0.4\textwidth}
\centering
\input{hvi_table}
\vspace{-2mm}
\caption{The HVI scale illustrates the hallucination tendencies exhibited by various LLMs.}
  \label{tab:hvi_spectrum}
\endminipage
\minipage{0.05\textwidth}
\vspace{-17mm}
\includegraphics[height=4.5cm]{img/hvi_arrow.pdf}
\endminipage
\vspace{-4.5mm}
\end{figure}
\end{comment}
\vspace{-1mm}
\section{Conclusion}
\vspace{-1mm}
The growing adoption and success of LLMs have been remarkable, yet they face a critical challenge: hallucination. While recent works have explored hallucination mitigation, automatic detection remains underexplored. To bridge this gap, we present $\mathcal{FACTOID}$, a dataset and benchmark for automatic hallucination detection. Our Factual Entailment technique has shown promising performance. We are committed to sharing all resources developed openly for further research.

%The LLMs suffer from many challenges one of them being hallucination. Additionally, there exists no standard metric to evaluate it automatically. We address this gap by introducing $\mathcal{FACTOID}$ -  an automated way to measure hallucination

\newpage

\newpage
\section{Discussion and Limitations} \label{sec:limitaions}
\vspace{-2mm}
\textbf{Discussion:} On June 14th, 2023, the European Parliament successfully passed its version of the EU AI Act \cite{euaiproposal}. Following this, many other countries began discussing their stance on the evolving realm of Generative AI. A primary agenda of policymaking is to protect citizens from political, digital, and physical security risks posed by Generative AI. While safeguarding against misuse is crucial, one of the biggest concerns among policymakers is the occurrence of unwanted errors by systems, such as hallucination (source: https://cetas.turing.ac.uk/publications/rapid-rise-generative-ai).

\textbf{Limitations:} The empirical findings indicate that classifying temporal issues poses the greatest challenge, as shown in Figure ~\ref{fig:fe_results}.  \cite{gurnee2023language} claimed that LLMs acquire linear representations of space and time across various scales, it is expected that LLM hold such information internally and can classify accordingly. Performance on temporal issue 66\% is not bad, but could be seen as a future direction to improve.

\section{Ethical Considerations}
Through our experiments, we have uncovered the susceptibility of LLMs to hallucination. 
While emphasizing the vulnerabilities of LLMs, our goal is to underscore their current limitations. However, it's crucial to address the potential misuse of our findings by malicious entities who might exploit AI-generated text for nefarious purposes, such as designing new adversarial attacks or creating fake news that is indistinguishable from human-written content. We strongly discourage such misuse and strongly advise against it.

\bibliography{factoid}

\begin{thebibliography}{60}
\expandafter\ifx\csname natexlab\endcsname\relax\def\natexlab#1{#1}\fi

\bibitem[{Almazrouei et~al.(2023)Almazrouei, Alobeidli, Alshamsi, Cappelli, Cojocaru, Debbah, Goffinet, Hesslow, Launay, Malartic et~al.}]{almazrouei2023falcon}
Ebtesam Almazrouei, Hamza Alobeidli, Abdulaziz Alshamsi, Alessandro Cappelli, Ruxandra Cojocaru, M{\'e}rouane Debbah, {\'E}tienne Goffinet, Daniel Hesslow, Julien Launay, Quentin Malartic, et~al. 2023.
\newblock The falcon series of open language models.
\newblock \emph{arXiv preprint arXiv:2311.16867}.

\bibitem[{Bayat et~al.(2023)Bayat, Qian, Han, Sang, Belyi, Khorshidi, Wu, Ilyas, and Li}]{bayat2023fleek}
Farima~Fatahi Bayat, Kun Qian, Benjamin Han, Yisi Sang, Anton Belyi, Samira Khorshidi, Fei Wu, Ihab~F. Ilyas, and Yunyao Li. 2023.
\newblock \href {http://arxiv.org/abs/2310.17119} {Fleek: Factual error detection and correction with evidence retrieved from external knowledge}.

\bibitem[{Beeching et~al.(2023)Beeching, Fourrier, Habib, Han, Lambert, Rajani, Sanseviero, Tunstall, and Wolf}]{open-llm-leaderboard}
Edward Beeching, Clémentine Fourrier, Nathan Habib, Sheon Han, Nathan Lambert, Nazneen Rajani, Omar Sanseviero, Lewis Tunstall, and Thomas Wolf. 2023.
\newblock Open llm leaderboard.
\newblock \url{https://huggingface.co/spaces/HuggingFaceH4/open_llm_leaderboard}.

\bibitem[{Bowman et~al.(2015)Bowman, Angeli, Potts, and Manning}]{bowman2015large}
Samuel~R Bowman, Gabor Angeli, Christopher Potts, and Christopher~D Manning. 2015.
\newblock \href {https://aclanthology.org/D15-1075/} {A large annotated corpus for learning natural language inference}.
\newblock \emph{arXiv preprint arXiv:1508.05326}.

\bibitem[{Brown et~al.(2020)Brown, Mann, Ryder, Subbiah, Kaplan, Dhariwal, Neelakantan, Shyam, Sastry, Askell et~al.}]{brown2020language}
Tom Brown, Benjamin Mann, Nick Ryder, Melanie Subbiah, Jared~D Kaplan, Prafulla Dhariwal, Arvind Neelakantan, Pranav Shyam, Girish Sastry, Amanda Askell, et~al. 2020.
\newblock \href {https://arxiv.org/abs/2005.14165} {Language models are few-shot learners}.
\newblock \emph{Advances in neural information processing systems}, 33:1877--1901.

\bibitem[{Chen et~al.(2023)Chen, Sikka, Cogswell, Ji, and Divakaran}]{chen2023dress}
Yangyi Chen, Karan Sikka, Michael Cogswell, Heng Ji, and Ajay Divakaran. 2023.
\newblock Dress: Instructing large vision-language models to align and interact with humans via natural language feedback.
\newblock \emph{arXiv preprint arXiv:2311.10081}.

\bibitem[{Cheng et~al.(2023)Cheng, Huang, Bi, Zhan, Liu, Wang, Sun, Wei, Deng, and Zhang}]{cheng2023uprise}
Daixuan Cheng, Shaohan Huang, Junyu Bi, Yuefeng Zhan, Jianfeng Liu, Yujing Wang, Hao Sun, Furu Wei, Denvy Deng, and Qi~Zhang. 2023.
\newblock \href {http://arxiv.org/abs/2303.08518} {Uprise: Universal prompt retrieval for improving zero-shot evaluation}.

\bibitem[{Chiang et~al.(2023)Chiang, Li, Lin, Sheng, Wu, Zhang, Zheng, Zhuang, Zhuang, Gonzalez, Stoica, and Xing}]{vicuna2023}
Wei-Lin Chiang, Zhuohan Li, Zi~Lin, Ying Sheng, Zhanghao Wu, Hao Zhang, Lianmin Zheng, Siyuan Zhuang, Yonghao Zhuang, Joseph~E. Gonzalez, Ion Stoica, and Eric~P. Xing. 2023.
\newblock \href {https://vicuna.lmsys.org} {Vicuna: An open-source chatbot impressing gpt-4 with 90\%* chatgpt quality}.

\bibitem[{Chuang et~al.(2023)Chuang, Xie, Luo, Kim, Glass, and He}]{chuang2023dola}
Yung-Sung Chuang, Yujia Xie, Hongyin Luo, Yoon Kim, James Glass, and Pengcheng He. 2023.
\newblock \href {http://arxiv.org/abs/2309.03883} {Dola: Decoding by contrasting layers improves factuality in large language models}.

\bibitem[{databricks(2023)}]{dolly}
databricks. 2023.
\newblock \href {https://www.databricks.com/blog/2023/04/12/dolly-first-open-commercially-viable-instruction-tuned-llm} {Dolly}.

\bibitem[{Deleu et~al.(2022)Deleu, Kanaa, Feng, Kerg, Bengio, Lajoie, and Bacon}]{DBLP:conf/iclr/DeleuKFKBLB22}
Tristan Deleu, David Kanaa, Leo Feng, Giancarlo Kerg, Yoshua Bengio, Guillaume Lajoie, and Pierre{-}Luc Bacon. 2022.
\newblock \href {https://openreview.net/forum?id=57PipS27Km} {Continuous-time meta-learning with forward mode differentiation}.
\newblock In \emph{The Tenth International Conference on Learning Representations, {ICLR} 2022, Virtual Event, April 25-29, 2022}. OpenReview.net.

\bibitem[{Elaraby et~al.(2023)Elaraby, Lu, Dunn, Zhang, Wang, Liu, Tian, Wang, and Wang}]{elaraby2023halo}
Mohamed Elaraby, Mengyin Lu, Jacob Dunn, Xueying Zhang, Yu~Wang, Shizhu Liu, Pingchuan Tian, Yuping Wang, and Yuxuan Wang. 2023.
\newblock \href {http://arxiv.org/abs/2308.11764} {Halo: Estimation and reduction of hallucinations in open-source weak large language models}.

\bibitem[{European-Parliament(2023)}]{euaiproposal}
European-Parliament. 2023.
\newblock \href {https://eur-lex.europa.eu/legal-content/EN/TXT/?uri=celex\%3A52021PC0206} {Proposal for a regulation of the european parliament and of the council laying down harmonised rules on artificial intelligence (artificial intelligence act) and amending certain union legislative acts}.

\bibitem[{Gao et~al.(2023)Gao, Dai, Pasupat, Chen, Chaganty, Fan, Zhao, Lao, Lee, Juan et~al.}]{gao2023rarr}
Luyu Gao, Zhuyun Dai, Panupong Pasupat, Anthony Chen, Arun~Tejasvi Chaganty, Yicheng Fan, Vincent Zhao, Ni~Lao, Hongrae Lee, Da-Cheng Juan, et~al. 2023.
\newblock Rarr: Researching and revising what language models say, using language models.
\newblock In \emph{Proceedings of the 61st Annual Meeting of the Association for Computational Linguistics (Volume 1: Long Papers)}, pages 16477--16508.

\bibitem[{Gu et~al.(2023)Gu, Dong, Wei, and Huang}]{gu2023knowledge}
Yuxian Gu, Li~Dong, Furu Wei, and Minlie Huang. 2023.
\newblock \href {http://arxiv.org/abs/2306.08543} {Knowledge distillation of large language models}.

\bibitem[{Gurnee and Tegmark(2023)}]{gurnee2023language}
Wes Gurnee and Max Tegmark. 2023.
\newblock \href {http://arxiv.org/abs/2310.02207} {Language models represent space and time}.

\bibitem[{Günther et~al.(2023)Günther, Milliken, Geuter, Mastrapas, Wang, and Xiao}]{günther2023jina}
Michael Günther, Louis Milliken, Jonathan Geuter, Georgios Mastrapas, Bo~Wang, and Han Xiao. 2023.
\newblock \href {http://arxiv.org/abs/2307.11224} {Jina embeddings: A novel set of high-performance sentence embedding models}.

\bibitem[{Jones et~al.(2023)Jones, Palangi, Simões, Chandrasekaran, Mukherjee, Mitra, Awadallah, and Kamar}]{jones2023teaching}
Erik Jones, Hamid Palangi, Clarisse Simões, Varun Chandrasekaran, Subhabrata Mukherjee, Arindam Mitra, Ahmed Awadallah, and Ece Kamar. 2023.
\newblock \href {http://arxiv.org/abs/2310.06827} {Teaching language models to hallucinate less with synthetic tasks}.

\bibitem[{Joshi et~al.(2020)Joshi, Chen, Liu, Weld, Zettlemoyer, and Levy}]{joshi-etal-2020-spanbert}
Mandar Joshi, Danqi Chen, Yinhan Liu, Daniel~S. Weld, Luke Zettlemoyer, and Omer Levy. 2020.
\newblock \href {https://doi.org/10.1162/tacl_a_00300} {{S}pan{BERT}: Improving pre-training by representing and predicting spans}.
\newblock \emph{Transactions of the Association for Computational Linguistics}, 8:64--77.

\bibitem[{Kang et~al.(2023)Kang, Ni, and Yao}]{kang2023ever}
Haoqiang Kang, Juntong Ni, and Huaxiu Yao. 2023.
\newblock \href {http://arxiv.org/abs/2311.09114} {Ever: Mitigating hallucination in large language models through real-time verification and rectification}.

\bibitem[{Ladhak et~al.(2023)Ladhak, Durmus, Suzgun, Zhang, Jurafsky, Mckeown, and Hashimoto}]{ladhak2023pre}
Faisal Ladhak, Esin Durmus, Mirac Suzgun, Tianyi Zhang, Dan Jurafsky, Kathleen Mckeown, and Tatsunori~B Hashimoto. 2023.
\newblock \href {https://aclanthology.org/2023.eacl-main.234.pdf} {When do pre-training biases propagate to downstream tasks? a case study in text summarization}.
\newblock In \emph{Proceedings of the 17th Conference of the European Chapter of the Association for Computational Linguistics}, pages 3198--3211.

\bibitem[{Lee et~al.(2022)Lee, Ping, Xu, Patwary, Fung, Shoeybi, and Catanzaro}]{NEURIPS2022_df438caa}
Nayeon Lee, Wei Ping, Peng Xu, Mostofa Patwary, Pascale~N Fung, Mohammad Shoeybi, and Bryan Catanzaro. 2022.
\newblock \href {https://proceedings.neurips.cc/paper_files/paper/2022/file/df438caa36714f69277daa92d608dd63-Paper-Conference.pdf} {Factuality enhanced language models for open-ended text generation}.
\newblock In \emph{Advances in Neural Information Processing Systems}, volume~35, pages 34586--34599. Curran Associates, Inc.

\bibitem[{Li et~al.(2023)Li, Patel, Vi{\'e}gas, Pfister, and Wattenberg}]{li2023inference}
Kenneth Li, Oam Patel, Fernanda Vi{\'e}gas, Hanspeter Pfister, and Martin Wattenberg. 2023.
\newblock Inference-time intervention: Eliciting truthful answers from a language model.
\newblock \emph{arXiv preprint arXiv:2306.03341}.

\bibitem[{Liu et~al.(2023)Liu, Xia, Wang, and Zhang}]{liu2023your}
Jiawei Liu, Chunqiu~Steven Xia, Yuyao Wang, and Lingming Zhang. 2023.
\newblock \href {https://arxiv.org/abs/2305.01210} {Is your code generated by chatgpt really correct? rigorous evaluation of large language models for code generation}.
\newblock \emph{arXiv preprint arXiv:2305.01210}.

\bibitem[{Liu et~al.(2019)Liu, Ott, Goyal, Du, Joshi, Chen, Levy, Lewis, Zettlemoyer, and Stoyanov}]{liu2019roberta}
Yinhan Liu, Myle Ott, Naman Goyal, Jingfei Du, Mandar Joshi, Danqi Chen, Omer Levy, Mike Lewis, Luke Zettlemoyer, and Veselin Stoyanov. 2019.
\newblock \href {https://arxiv.org/abs/1907.11692} {Roberta: A robustly optimized bert pretraining approach}.
\newblock \emph{arXiv preprint arXiv:1907.11692}.

\bibitem[{Maynez et~al.(2020)Maynez, Narayan, Bohnet, and McDonald}]{maynez-etal-2020-faithfulness}
Joshua Maynez, Shashi Narayan, Bernd Bohnet, and Ryan McDonald. 2020.
\newblock \href {https://doi.org/10.18653/v1/2020.acl-main.173} {On faithfulness and factuality in abstractive summarization}.
\newblock In \emph{Proceedings of the 58th Annual Meeting of the Association for Computational Linguistics}, pages 1906--1919, Online. Association for Computational Linguistics.

\bibitem[{Midjourney(2022)}]{midjourney}
Midjourney. 2022.
\newblock \href {https://www.midjourney.com} {https://www.midjourney.com}.

\bibitem[{Mikolov et~al.(2013)Mikolov, Sutskever, Chen, Corrado, and Dean}]{mikolov2013distributed}
Tomas Mikolov, Ilya Sutskever, Kai Chen, Greg~S Corrado, and Jeff Dean. 2013.
\newblock Distributed representations of words and phrases and their compositionality.
\newblock \emph{Advances in neural information processing systems}, 26.

\bibitem[{Muennighoff et~al.(2022)Muennighoff, Tazi, Magne, and Reimers}]{muennighoff2022mteb}
Niklas Muennighoff, Nouamane Tazi, Lo{\"\i}c Magne, and Nils Reimers. 2022.
\newblock \href {https://doi.org/10.48550/ARXIV.2210.07316} {Mteb: Massive text embedding benchmark}.
\newblock \emph{arXiv preprint arXiv:2210.07316}.

\bibitem[{M{\"u}ndler et~al.(2023)M{\"u}ndler, He, Jenko, and Vechev}]{mundler2023self}
Niels M{\"u}ndler, Jingxuan He, Slobodan Jenko, and Martin Vechev. 2023.
\newblock Self-contradictory hallucinations of large language models: Evaluation, detection and mitigation.
\newblock \emph{arXiv preprint arXiv:2305.15852}.

\bibitem[{OpenAI(2022)}]{ChatGPT}
OpenAI. 2022.
\newblock \href {https://openai.com/blog/chatgpt} {Introducing chatgpt}.

\bibitem[{OpenAI(2023)}]{openai2023gpt4}
OpenAI. 2023.
\newblock \href {http://arxiv.org/abs/2303.08774} {Gpt-4 technical report}.

\bibitem[{Papineni et~al.(2002)Papineni, Roukos, Ward, and Zhu}]{papineni2002bleu}
Kishore Papineni, Salim Roukos, Todd Ward, and Wei-Jing Zhu. 2002.
\newblock Bleu: a method for automatic evaluation of machine translation.
\newblock In \emph{Proceedings of the 40th annual meeting of the Association for Computational Linguistics}, pages 311--318.

\bibitem[{Peng et~al.(2023)Peng, Galley, He, Cheng, Xie, Hu, Huang, Liden, Yu, Chen et~al.}]{peng2023check}
Baolin Peng, Michel Galley, Pengcheng He, Hao Cheng, Yujia Xie, Yu~Hu, Qiuyuan Huang, Lars Liden, Zhou Yu, Weizhu Chen, et~al. 2023.
\newblock Check your facts and try again: Improving large language models with external knowledge and automated feedback.
\newblock \emph{arXiv preprint arXiv:2302.12813}.

\bibitem[{Qiu et~al.(2023{\natexlab{a}})Qiu, Embar, Cohen, and Han}]{qiu2023think}
Yifu Qiu, Varun Embar, Shay~B Cohen, and Benjamin Han. 2023{\natexlab{a}}.
\newblock Think while you write: Hypothesis verification promotes faithful knowledge-to-text generation.
\newblock \emph{arXiv preprint arXiv:2311.09467}.

\bibitem[{Qiu et~al.(2023{\natexlab{b}})Qiu, Ziser, Korhonen, Ponti, and Cohen}]{qiu2023detecting}
Yifu Qiu, Yftah Ziser, Anna Korhonen, Edoardo~M Ponti, and Shay~B Cohen. 2023{\natexlab{b}}.
\newblock Detecting and mitigating hallucinations in multilingual summarisation.
\newblock \emph{arXiv preprint arXiv:2305.13632}.

\bibitem[{Radford et~al.(2019)Radford, Wu, Child, Luan, Amodei, Sutskever et~al.}]{radford2019language}
Alec Radford, Jeffrey Wu, Rewon Child, David Luan, Dario Amodei, Ilya Sutskever, et~al. 2019.
\newblock \href {https://d4mucfpksywv.cloudfront.net/better-language-models/language_models_are_unsupervised_multitask_learners.pdf} {Language models are unsupervised multitask learners}.
\newblock \emph{OpenAI blog}, 1(8):9.

\bibitem[{Raffel et~al.(2020)Raffel, Shazeer, Roberts, Lee, Narang, Matena, Zhou, Li, and Liu}]{raffel2020exploring}
Colin Raffel, Noam Shazeer, Adam Roberts, Katherine Lee, Sharan Narang, Michael Matena, Yanqi Zhou, Wei Li, and Peter~J Liu. 2020.
\newblock \href {https://www.jmlr.org/papers/volume21/20-074/20-074.pdf} {Exploring the limits of transfer learning with a unified text-to-text transformer}.
\newblock \emph{The Journal of Machine Learning Research}, 21(1):5485--5551.

\bibitem[{Ramesh et~al.(2022)Ramesh, Dhariwal, Nichol, Chu, and Chen}]{ramesh2022hierarchical}
Aditya Ramesh, Prafulla Dhariwal, Alex Nichol, Casey Chu, and Mark Chen. 2022.
\newblock \href {https://arxiv.org/abs/2204.06125} {Hierarchical text-conditional image generation with clip latents}.
\newblock \emph{arXiv preprint arXiv:2204.06125}.

\bibitem[{Ramesh et~al.(2021)Ramesh, Pavlov, Goh, Gray, Voss, Radford, Chen, and Sutskever}]{ramesh2021zero}
Aditya Ramesh, Mikhail Pavlov, Gabriel Goh, Scott Gray, Chelsea Voss, Alec Radford, Mark Chen, and Ilya Sutskever. 2021.
\newblock \href {https://arxiv.org/abs/2102.12092} {Zero-shot text-to-image generation}.
\newblock In \emph{International Conference on Machine Learning}, pages 8821--8831. PMLR.

\bibitem[{Raunak et~al.(2021)Raunak, Menezes, and Junczys-Dowmunt}]{raunak-etal-2021-curious}
Vikas Raunak, Arul Menezes, and Marcin Junczys-Dowmunt. 2021.
\newblock \href {https://doi.org/10.18653/v1/2021.naacl-main.92} {The curious case of hallucinations in neural machine translation}.
\newblock In \emph{Proceedings of the 2021 Conference of the North American Chapter of the Association for Computational Linguistics: Human Language Technologies}, pages 1172--1183, Online. Association for Computational Linguistics.

\bibitem[{Rawte et~al.(2023)Rawte, Chakraborty, Pathak, Sarkar, Tonmoy, Chadha, Sheth, and Das}]{rawte2023troubling}
Vipula Rawte, Swagata Chakraborty, Agnibh Pathak, Anubhav Sarkar, SM~Tonmoy, Aman Chadha, Amit~P Sheth, and Amitava Das. 2023.
\newblock The troubling emergence of hallucination in large language models--an extensive definition, quantification, and prescriptive remediations.
\newblock \emph{arXiv preprint arXiv:2310.04988}.

\bibitem[{Rombach et~al.(2022)Rombach, Blattmann, Lorenz, Esser, and Ommer}]{rombach2022high}
Robin Rombach, Andreas Blattmann, Dominik Lorenz, Patrick Esser, and Bj{\"o}rn Ommer. 2022.
\newblock \href {https://arxiv.org/abs/2112.10752} {High-resolution image synthesis with latent diffusion models}.
\newblock In \emph{Proceedings of the IEEE/CVF Conference on Computer Vision and Pattern Recognition}, pages 10684--10695.

\bibitem[{Scao et~al.(2022)Scao, Fan, Akiki, Pavlick, Ili{\'c}, Hesslow, Castagn{\'e}, Luccioni, Yvon, Gall{\'e} et~al.}]{scao2022bloom}
Teven~Le Scao, Angela Fan, Christopher Akiki, Ellie Pavlick, Suzana Ili{\'c}, Daniel Hesslow, Roman Castagn{\'e}, Alexandra~Sasha Luccioni, Fran{\c{c}}ois Yvon, Matthias Gall{\'e}, et~al. 2022.
\newblock \href {https://arxiv.org/abs/2211.05100} {Bloom: A 176b-parameter open-access multilingual language model}.
\newblock \emph{arXiv preprint arXiv:2211.05100}.

\bibitem[{Si et~al.(2022)Si, Gan, Yang, Wang, Wang, Boyd-Graber, and Wang}]{si2022prompting}
Chenglei Si, Zhe Gan, Zhengyuan Yang, Shuohang Wang, Jianfeng Wang, Jordan Boyd-Graber, and Lijuan Wang. 2022.
\newblock Prompting gpt-3 to be reliable.
\newblock \emph{arXiv preprint arXiv:2210.09150}.

\bibitem[{Su et~al.(2022)Su, Lu, Pan, Murtadha, Wen, and Liu}]{su2022roformer}
Jianlin Su, Yu~Lu, Shengfeng Pan, Ahmed Murtadha, Bo~Wen, and Yunfeng Liu. 2022.
\newblock \href {http://arxiv.org/abs/2104.09864} {Roformer: Enhanced transformer with rotary position embedding}.

\bibitem[{Sudre et~al.(2017)Sudre, Li, Vercauteren, Ourselin, and Jorge~Cardoso}]{Sudre_2017}
Carole~H. Sudre, Wenqi Li, Tom Vercauteren, Sebastien Ourselin, and M.~Jorge~Cardoso. 2017.
\newblock \href {https://doi.org/10.1007/978-3-319-67558-9_28} {\emph{Generalised Dice Overlap as a Deep Learning Loss Function for Highly Unbalanced Segmentations}}, page 240–248. Springer International Publishing.

\bibitem[{Taori et~al.(2023)Taori, Gulrajani, Zhang, Dubois, Li, Guestrin, Liang, and Hashimoto}]{alpaca}
Rohan Taori, Ishaan Gulrajani, Tianyi Zhang, Yann Dubois, Xuechen Li, Carlos Guestrin, Percy Liang, and Tatsunori~B. Hashimoto. 2023.
\newblock \href {https://crfm.stanford.edu/2023/03/13/alpaca.html} {Stanford alpaca: An instruction-following llama model}.
\newblock \url{https://github.com/tatsu-lab/stanford_alpaca}.

\bibitem[{Tian et~al.(2023)Tian, Mitchell, Yao, Manning, and Finn}]{tian2023finetuning}
Katherine Tian, Eric Mitchell, Huaxiu Yao, Christopher~D. Manning, and Chelsea Finn. 2023.
\newblock \href {http://arxiv.org/abs/2311.08401} {Fine-tuning language models for factuality}.

\bibitem[{Touvron et~al.(2023)Touvron, Martin, Stone, Albert, Almahairi, Babaei, Bashlykov, Batra, Bhargava, Bhosale et~al.}]{touvron2023llama}
Hugo Touvron, Louis Martin, Kevin Stone, Peter Albert, Amjad Almahairi, Yasmine Babaei, Nikolay Bashlykov, Soumya Batra, Prajjwal Bhargava, Shruti Bhosale, et~al. 2023.
\newblock Llama 2: Open foundation and fine-tuned chat models.
\newblock \emph{arXiv preprint arXiv:2307.09288}.

\bibitem[{Vu et~al.(2023)Vu, Iyyer, Wang, Constant, Wei, Wei, Tar, Sung, Zhou, Le, and Luong}]{vu2023freshllms}
Tu~Vu, Mohit Iyyer, Xuezhi Wang, Noah Constant, Jerry Wei, Jason Wei, Chris Tar, Yun-Hsuan Sung, Denny Zhou, Quoc Le, and Thang Luong. 2023.
\newblock \href {http://arxiv.org/abs/2310.03214} {Freshllms: Refreshing large language models with search engine augmentation}.

\bibitem[{Wagner and Fischer(1974)}]{wagner1974string}
Robert~A Wagner and Michael~J Fischer. 1974.
\newblock The string-to-string correction problem.
\newblock \emph{Journal of the ACM (JACM)}, 21(1):168--173.

\bibitem[{Wang et~al.(2023)Wang, Panda, Karlinsky, Feris, Sun, and Kim}]{wang2023multitask}
Zhen Wang, Rameswar Panda, Leonid Karlinsky, Rogerio Feris, Huan Sun, and Yoon Kim. 2023.
\newblock \href {https://openreview.net/forum?id=Nk2pDtuhTq} {Multitask prompt tuning enables parameter-efficient transfer learning}.
\newblock In \emph{The Eleventh International Conference on Learning Representations}.

\bibitem[{Wikipedia\_min\_max()}]{Normalization-min}
Wikipedia\_min\_max.
\newblock \href {https://en.wikipedia.org/wiki/Normalization_(statistics)} {Normalization}.

\bibitem[{Wikipedia\_zscore()}]{Normalization-z}
Wikipedia\_zscore.
\newblock \href {https://en.wikipedia.org/wiki/Normalization_(statistics)} {Normalization}.

\bibitem[{Williams et~al.(2018)Williams, Nangia, and Bowman}]{williams-etal-2018-broad}
Adina Williams, Nikita Nangia, and Samuel Bowman. 2018.
\newblock \href {https://doi.org/10.18653/v1/N18-1101} {A broad-coverage challenge corpus for sentence understanding through inference}.
\newblock In \emph{Proceedings of the 2018 Conference of the North {A}merican Chapter of the Association for Computational Linguistics: Human Language Technologies, Volume 1 (Long Papers)}, pages 1112--1122, New Orleans, Louisiana. Association for Computational Linguistics.

\bibitem[{Yang et~al.(2019)Yang, Dai, Yang, Carbonell, Salakhutdinov, and Le}]{yang2019xlnet}
Zhilin Yang, Zihang Dai, Yiming Yang, Jaime Carbonell, Russ~R Salakhutdinov, and Quoc~V Le. 2019.
\newblock \href {https://proceedings.neurips.cc/paper/2019/file/dc6a7e655d7e5840e66733e9ee67cc69-Paper.pdf} {Xlnet: Generalized autoregressive pretraining for language understanding}.
\newblock \emph{Advances in neural information processing systems}, 32.

\bibitem[{Yoon et~al.(2022)Yoon, Yoon, Yoon, Kim, and Yoo}]{yoon-etal-2022-information}
Sunjae Yoon, Eunseop Yoon, Hee~Suk Yoon, Junyeong Kim, and Chang Yoo. 2022.
\newblock \href {https://doi.org/10.18653/v1/2022.emnlp-main.280} {Information-theoretic text hallucination reduction for video-grounded dialogue}.
\newblock In \emph{Proceedings of the 2022 Conference on Empirical Methods in Natural Language Processing}, pages 4182--4193, Abu Dhabi, United Arab Emirates. Association for Computational Linguistics.

\bibitem[{Zhang et~al.(2022)Zhang, Roller, Goyal, Artetxe, Chen, Chen, Dewan, Diab, Li, Lin, Mihaylov, Ott, Shleifer, Shuster, Simig, Koura, Sridhar, Wang, and Zettlemoyer}]{zhang2022opt}
Susan Zhang, Stephen Roller, Naman Goyal, Mikel Artetxe, Moya Chen, Shuohui Chen, Christopher Dewan, Mona Diab, Xian Li, Xi~Victoria Lin, Todor Mihaylov, Myle Ott, Sam Shleifer, Kurt Shuster, Daniel Simig, Punit~Singh Koura, Anjali Sridhar, Tianlu Wang, and Luke Zettlemoyer. 2022.
\newblock \href {http://arxiv.org/abs/2205.01068} {Opt: Open pre-trained transformer language models}.

\bibitem[{Ziegler et~al.(2019)Ziegler, Stiennon, Wu, Brown, Radford, Amodei, Christiano, and Irving}]{DBLP:journals/corr/abs-1909-08593}
Daniel~M. Ziegler, Nisan Stiennon, Jeffrey Wu, Tom~B. Brown, Alec Radford, Dario Amodei, Paul~F. Christiano, and Geoffrey Irving. 2019.
\newblock \href {http://arxiv.org/abs/1909.08593} {Fine-tuning language models from human preferences}.
\newblock \emph{CoRR}, abs/1909.08593.

\end{thebibliography}
\bibliographystyle{acl_natbib}

\newpage
\newpage
\onecolumn

\section*{Frequently Asked Questions (FAQs)}\label{sec:FAQs}

\begin{itemize}[leftmargin=*,nolistsep]
%[leftmargin=0mm]
%\setlength\itemsep{0em}
    \item[\ding{93}] {\fontfamily{lmss} \selectfont \textbf{This study explores the unintended, negative aspects of hallucination; how about the useful effects that arise as a result of hallucination?}}
    \vspace{0mm}
    \begin{description}
    \item[\ding{224}] While hallucinating has beneficiary effects in some computer vision use cases, where a generative vision model could perform in-painting of an occluded content in an image or generate an image of a scenario it hasn't seen in its training set (for example, a generated image corresponding to the prompt, ``water on Mars''), but it is usually undesirable in the context of the text. The downstream impact as a result of the model's is exacerbated by the fact that there is a lack of a programmatic method in the research community to distinguish the hallucinated vs. factually correct output. For this reason, this study focuses on characterizing the problem of hallucination particularly in the context of text.
    \end{description}
\vspace{2mm}
    \item[\ding{93}] {\fontfamily{lmss} \selectfont \textbf{Why do you select those 15 large language models?}}
    \vspace{0mm}
    \begin{description}
    \item[\ding{224}] We want to select several language models with varying parameter sizes for our experiments - ranging from large to small. Hence, the above chosen 14 models consist of large models like GPT-3 and smaller ones like T5 and T0.
    \end{description}
\vspace{2mm}
%    \item[\ding{93}] {\fontfamily{lmss} \selectfont \textbf{Why would extrinsic hallucination be riskier?}}
%    \vspace{-2mm}
%    \begin{description}
%    \item[\ding{224}] According to the ``extrinsic hallucination'' definition, this kind of hallucination does not have any way to verify it from the source prompt. Hence, it is likely to be more harmful than the intrinsic ones.
%    \end{description}

%    \item[\ding{93}] {\fontfamily{lmss} \selectfont \textbf{What is the purpose of constructing Factual Mirage and Silver Lining hallucination data?}}
%    \vspace{-2mm}
%    \begin{description}
%    \item[\ding{224}] We want to show that hallucinations can happen in both cases, factually correct and incorrect prompts. Hence, in this paper, we construct an exhaustive dataset called 
%\end{description}

%    \item[\ding{93}] {\fontfamily{lmss} \selectfont \textbf{Why do you select high-entropy points for mitigation techniques?}}
%    \vspace{-2mm}
%    \begin{description}
%    \item[\ding{224}] High entropy points are more uncertain points in the context of text generation and hence, more likely places where the LLM hallucinates. Hence, our mitigation approach works by detecting and replacing such high entropy points.
%    \end{description}

    \item[\ding{93}] {\fontfamily{lmss} \selectfont \textbf{Why would HVI be a better hallucination evaluation metric for the LLMs (as compared to the existing ones like accuracy, precision, recall, F1, etc.)?}}
    \vspace{0mm}
    \begin{description}
    \item[\ding{224}] Although the commonly used evaluation metrics like accuracy, precision, etc. can be used for downstream tasks, HVI can be more specifically used to determine the LLMs' hallucination tendency. HVI will serve as a uniform hallucination score for all the present and future LLMs.
    \end{description}

%    \item[\ding{93}] {\fontfamily{lmss} \selectfont \textbf{What are the insights on using black-box vs. gray-box models for mitigation hallucinations?}}
%    \vspace{-2mm}
%    \begin{description}
%    \item[\ding{224}] Both black-box and gray-box models have their own advantages and disadvantages in terms of reducing hallucinations. Therefore, the choice of the appropriate method to minimize hallucination would be LLM- and task-dependent.
%    \end{description}

\end{itemize}    
\newpage
\appendix

\section{Appendix}
\label{sec:appendix}

%\vspace{-6mm}
%\input{hvi_equation}

\begin{comment}
\appendix
\renewcommand{\thesubsection}{\Alph{section}.\arabic{subsection}}
\renewcommand{\thesection}{\Alph{section}}
\setcounter{section}{0}
\end{comment}

% \def\thesection{\alpha{section}}
% \def\thesubsection{\thesection.\arabic{subsection}}
%\section*{Appendix}\label{sec:appendix}

This section provides supplementary material in the form of additional examples, implementation details, etc. to bolster the reader's understanding of the concepts presented in this work.

\section{Annotation Process, and agreement}
\label{subsec:annotation}

In the initial in-house annotation phase, crowdsourcing platforms are acknowledged for their speed and cost-effectiveness in annotation tasks. Nevertheless, it's crucial to acknowledge that they may introduce noise or inaccuracies. To address this, prior to engaging crowdsourcing services, we conducted an in-house annotation process involving 1,000 samples.

\section{Paraphrasing}  \label{sec:para}
\textbf{Coverage - Quantity of Significant Paraphrase Generations:} Our aim is to create up to $5$ paraphrases for each claim. Following the generation of claims, we employ the Minimum Edit Distance (MED) \cite{wagner1974string}—measured in words, not alphabets. If the MED exceeds $\pm2$ for any paraphrase candidate (e.g., $c-p_1^c$) with the claim, we include that paraphrase; otherwise, we discard it. We assess all three models based on their ability to generate a substantial number of paraphrases.

\textbf{Correctness - Accuracy in Paraphrase Generations:} Post the initial filtration, we conduct pairwise entailment, retaining paraphrase candidates marked as entailed by \cite{liu2019roberta} (Roberta Large), a state-of-the-art model trained on SNLI \cite{bowman2015large}.

\textbf{Diversity - Linguistic Variety in Paraphrase Generations:} Our focus is on selecting a model capable of producing linguistically diverse paraphrases. We assess dissimilarities among generated paraphrase claims—for instance, $c-p_n^c$, $p_1^c-p_n^c$, $p_2^c-p_n^c$, and so on. This process is repeated for all paraphrases, averaging out the dissimilarity score. Lacking a specific dissimilarity metric, we use the inverse of the BLEU score \cite{papineni2002bleu}. This provides insight into how linguistic diversity is achieved by a given model. Our experiments reveal that \texttt{gpt-3.5-turbo-0301} performs the best, as reported in the table. Additionally, we prioritize a model that maximizes linguistic variations, and \texttt{gpt-3.5-turbo-0301} excels in this aspect. A plot illustrating diversity versus all chosen models is presented in ~\cref{fig:parr}.

\section{FACTOID dataset creation}  \label{sec:dat}

The process for creating the synthetic dataset is given in \cref{alg:data},

\begin{algorithm}
\caption{Creating \emph{positive-negative} samples}\label{alg:data}
\begin{algorithmic}

\For{each factually correct prompt $f$}
    \State find the named entities causing hallucination 
    \State find top-$5$ similar entities in the vector space using \emph{word2vec} $\{s_1, s_2, s_3, s_4, s_5\}$
    \For{each similar entity $s$}
    \State replace the original entity with a similar entity
    \State generate $5$ paraphrases $\{p_1, p_2, p_3, p_4, p_5\}$
\EndFor
\EndFor
\end{algorithmic}
\end{algorithm}

\begin{figure*}
    \centering
    \includegraphics[width=\textwidth]{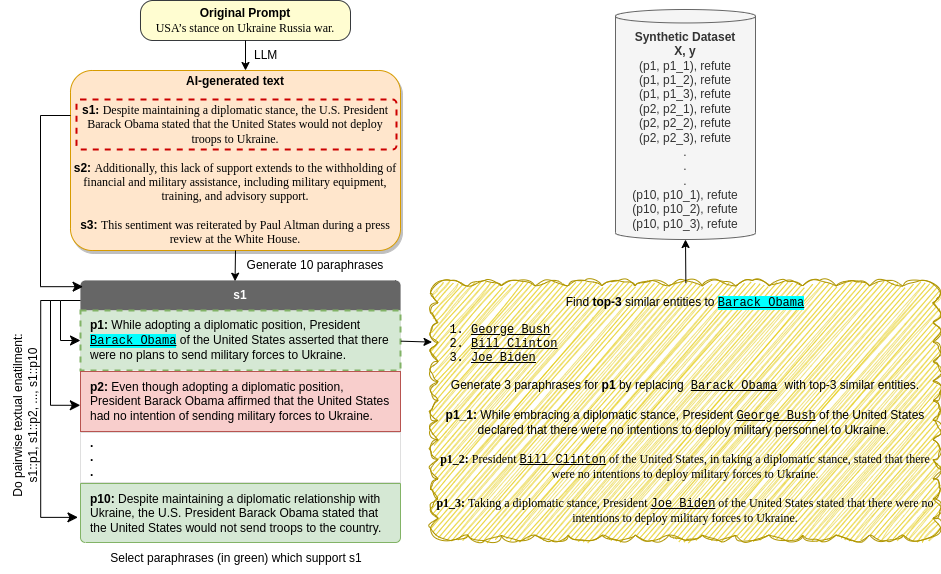}
    \caption{Process to generate synthetic data.}
    \label{fig:acorn-para}
\end{figure*}

\begin{figure}[!ht]
    \centering
    \includegraphics[width=0.5\textwidth]{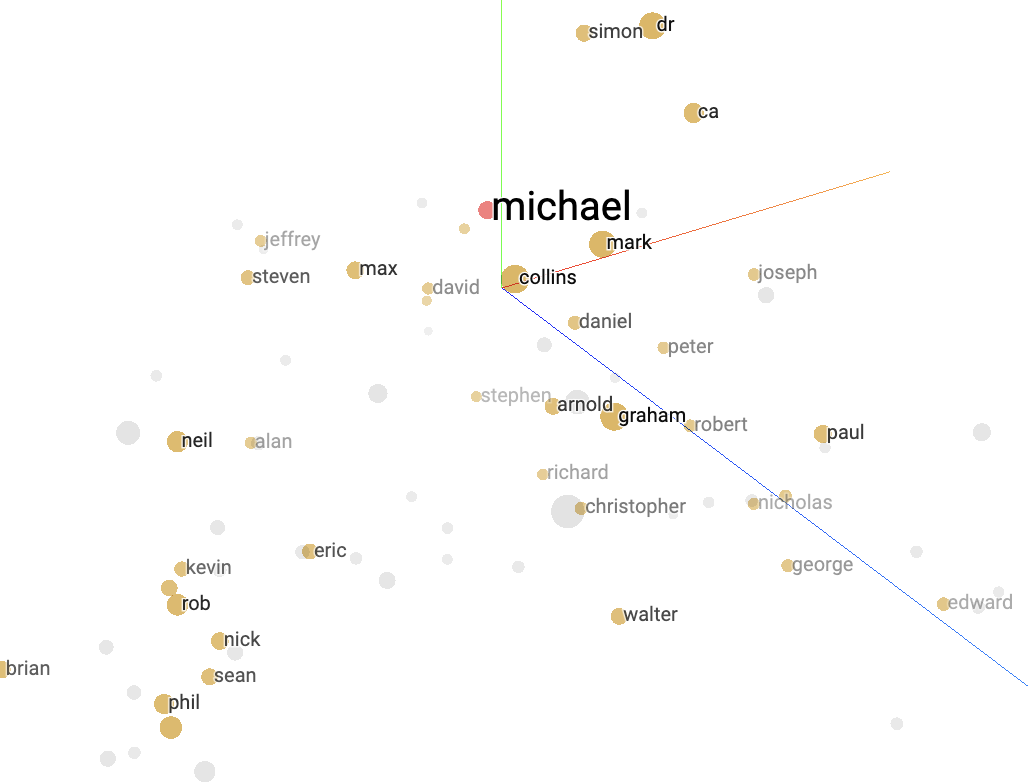}
    \caption{similar person names}
    \label{fig:fe}
\end{figure}

\section{Longer embedding}
Long-text embeddings are crafted to represent textual content and grasp the semantic essence of lengthy passages. In contrast to conventional embeddings for shorter texts that might face challenges in preserving context, longer text embeddings shine in capturing information from detailed articles, expansive books, or extensive documents. Defined by higher dimensions, usually spanning from 768 to 4096, they enable a nuanced understanding and the capture of relationships within extended textual contexts.

\subsection{Long-Text High-Dimensional Embeddings}

In the realm of NLP, the advent of long-text embeddings marks a pivotal evolution from traditional, shorter embeddings, addressing critical limitations and broadening the application spectrum. Long-text embeddings, typically high dimensional ranging from 768 to 4096 dimensions, have emerged as a crucial innovation, primarily for their adeptness at encapsulating the semantics of extensive texts, ranging from detailed articles to comprehensive books. This capability significantly enhances document-level understanding, allowing for a more nuanced grasp of themes, narrative structures, argumentative patterns, etc. Moreover, the ability to process and analyze texts in their entirety without truncation reduces information loss, ensuring that vital context and intricate details are preserved. Long-text embeddings excel in capturing long-distance relationships and dependencies within texts, a feature that is instrumental for tasks requiring deep contextual interpretation such as question answering and textual entailment. Furthermore, these embeddings facilitate complex analyses, including thematic development, stylistic evolution, and sentiment tracking across lengthy documents, opening new avenues in literary analysis, historical research, and more. The shift towards longer text embeddings thus represents a significant leap forward in NLP, enabling more accurate, comprehensive, and sophisticated text processing and analysis, thereby overcoming the constraints posed by shorter embeddings and unlocking new potentials in understanding and leveraging large-scale textual data. 
This deep-rooted understanding offered by long-text embeddings is particularly beneficial for tasks that require a holistic understanding of the broader context, coupled with a nuanced understanding of the immediate topic at hand, to infer factual irregularities and thus detect hallucinations. Using the MTEB Leaderboard \cite{muennighoff2022mteb}, we identified the top-performing long-text embedding models as of this writing, with a max-token limit ranging from 8K to 32K.

% , and selected the following for our experimentation: (i) \texttt{SFR-Embedding-Mistral} \cite{sfr}: Trained by Salesforce Research on top of \texttt{E5-mistral-7b-instruct} and \texttt{Mistral-7B-v0.1}, offers 4096-dimensional embeddings over 32K tokens; (ii) \texttt{e5-mistral-7b-instruct} \cite{wang2023improving}: Initialized from \texttt{Mistral-7B-v0.1} and fine-tuned on a mixture of multilingual datasets, offers 4096-dimensional embeddings over 32K tokens; (iii) \texttt{nomic-embed-text-v1} \cite{nussbaum2024nomic}: Trained by Nomic using a long-context BERT model with a two-stage process involving an unsupervised contrastive stage based on weakly related text pairs followed by higher quality labeled datasets, offers 768-dimensional embeddings over 8K tokens; (iv) \texttt{text-embedding-3-large} \cite{oai}: The next generation embedding model trained by OpenAI, offers 3072-dimensional embeddings over 8K tokens. 

The list of sentences is below:

\paragraph{sent1:}  "The sun sets behind the mountains, casting a warm glow across the landscape. The sky transforms into a canvas of vibrant hues, from fiery oranges to soft purples. The air becomes cooler as twilight descends upon the earth. Nature's evening symphony begins, with the chirping of crickets and the rustle of leaves in the gentle breeze. As night falls, the world settles into a peaceful slumber, awaiting the dawn of a new day.
    
\paragraph{sent2:}  "As the sun dips beneath the silhouette of the mountains, its departing rays blanket the land with a comforting warmth, creating a picturesque scene. Gradually, the sky undergoes a breathtaking transformation, transitioning from the blazing brilliance of oranges to the soothing tones of purples, creating a mesmerizing spectacle overhead. With the fading light, a gentle coolness pervades the atmosphere, signaling the onset of twilight, a time when the earth enters a state of tranquil transition. Nature, in its evening rituals, orchestrates a harmonious symphony, with the melodious chirping of crickets and the gentle rustling of leaves accompanying the fading daylight. And so, as the darkness of night descends, the world surrenders to a serene slumber, patiently awaiting the emergence of a new dawn, heralding the promise of another day."
    
\paragraph{sent3:}  "Behind the rugged peaks, the sun gracefully retreats, suffusing the landscape with a radiant warmth that caresses every contour of the earth. Across the vast expanse, the heavens burst into an array of vibrant colors, from the fiery embrace of oranges to the tranquil embrace of purples, painting a captivating tableau above. As daylight wanes, a gentle chill creeps into the air, heralding the arrival of twilight, a transitional phase where the world pauses to catch its breath. Nature, in its evening chorus, serenades the fading light with the rhythmic chirping of crickets and the soft whispers of leaves dancing in the breeze. And so, with the advent of night, the world succumbs to a tranquil slumber, embracing the promise of renewal with each passing moment until the dawn of a new day breaks upon the horizon."
    
\paragraph{sent4:}  "The descent of the sun beyond the jagged peaks casts a golden glow upon the land, enveloping it in a serene embrace. Across the vast expanse of the sky, a kaleidoscope of colors emerges, transitioning from the fiery intensity of oranges to the gentle hues of purples and pinks, creating a breathtaking panorama. With the fading light, a sense of calmness descends, as the air grows cooler and the world prepares for the arrival of twilight. Nature, in its evening symphony, orchestrates a melodious chorus, with the chirping of crickets and the rustling of leaves providing the soundtrack to the fading day. And so, as night falls, the world settles into a tranquil slumber, eagerly anticipating the promise of a new beginning with the break of dawn."
    
\paragraph{sent5:}  "Behind the majestic peaks, the sun bids adieu, casting a warm glow that envelops the landscape in a comforting embrace. The sky transforms into a canvas of breathtaking beauty, with hues ranging from the fiery brilliance of oranges to the soft pastels of purples and pinks, creating a mesmerizing display. As daylight fades, a gentle coolness fills the air, signaling the arrival of twilight, a magical time when the earth transitions into a state of serene tranquility. Nature, in its nightly ritual, comes alive with the chirping of crickets and the gentle rustling of leaves, as if bidding farewell to the departing day. And so, as darkness descends, the world settles into a peaceful slumber, eagerly awaiting the dawn of a new day and the promise it brings."
    
\paragraph{sent6:} "As the sun dips below the horizon, its fading rays cast a golden glow upon the land, imbuing it with a sense of warmth and serenity. Above, the sky transforms into a breathtaking tapestry of colors, with vibrant oranges giving way to soft purples and pinks, painting a scene of unparalleled beauty. With the onset of twilight, the air grows cooler, enveloping the world in a gentle embrace as it prepares for the night ahead. Nature, in its nightly symphony, fills the air with the soothing sounds of crickets chirping and leaves rustling, a melodic accompaniment to the fading light. And so, as night falls, the world settles into a peaceful slumber, eagerly anticipating the dawn of a new day and the endless possibilities it brings."

\section{Details of performance of FE}
%\adil{text...}
% \usepackage{booktabs}
% \usepackage{graphicx}
\begin{table}[h!]
\centering
\resizebox{\columnwidth}{!}{%
\begin{tabular}{@{}lrrrrr@{}}
\toprule
\textbf{Entailment technique/ Hallucination Type} &
  \multicolumn{1}{l}{\textbf{Imaginary Figure}} &
  \multicolumn{1}{l}{\textbf{Place}} &
  \multicolumn{1}{l}{\textbf{Bothersome Number}} &
  \multicolumn{1}{l}{\textbf{Temporal Issue}} &
  \multicolumn{1}{l}{\textbf{Avg.}} \\ \midrule
\textbf{Baseline (Traditional entailment)} &
  0.44 &
  0.49 &
  0.23 &
  0.12 &
  0.32 \\
\textbf{Factual entailment} &
  0.69 &
  0.71 &
  0.67 &
  0.59 &
  0.665 \\ \bottomrule
\end{tabular}%
}
\caption{}
\label{tab:my-table}
\end{table}

\end{document}